\documentclass[letterpaper,UKenglish,cleveref,autoref,thm-restate,numberwithinsect]{lipics-v2021}
\usepackage[ruled,linesnumbered]{algorithm2e}
\usepackage{booktabs,dsfont}

\SetKwInput{KwInput}{Input}
\SetKwInput{KwOutput}{Output}
\SetKwComment{Comment}{/* }{ */}
\newcommand{\diam}{\mathrm{diam}}
\newcommand{\indicator}{\mathds{1}}
\usepackage{float}

\allowdisplaybreaks

\usepackage{amsmath,amsfonts,bm}

\def\eqref#1{(\ref{#1})}

\def\ceil#1{\lceil #1 \rceil}
\def\floor#1{\lfloor #1 \rfloor}
\def\1{\bm{1}}

\def\vc{{\bm{c}}}

\DeclareMathAlphabet{\mathsfit}{\encodingdefault}{\sfdefault}{m}{sl}
\SetMathAlphabet{\mathsfit}{bold}{\encodingdefault}{\sfdefault}{bx}{n}

\def\gC{{\mathcal{C}}}

\def\gE{{\mathcal{E}}}

\def\gG{{\mathcal{G}}}
\def\gH{{\mathcal{H}}}

\def\gK{{\mathcal{K}}}

\def\gM{{\mathcal{M}}}
\def\gN{{\mathcal{N}}}

\def\gP{{\mathcal{P}}}

\def\gV{{\mathcal{V}}}

\def\sN{{\mathbb{N}}}

\def\sP{{\mathbb{P}}}

\def\sR{{\mathbb{R}}}

\newcommand{\E}{\mathbb{E}}

\newcommand{\KL}{D_{\mathrm{KL}}}

\DeclareMathOperator*{\argmax}{arg\,max}

\title{Flooding with Absorption: An Efficient Protocol for Heterogeneous Bandits over Complex Networks} 

\titlerunning{Flooding with Absorption} 
\author{Junghyun Lee}{Kim Jaechul Graduate School of AI, KAIST, Seoul, Republic of Korea \and \url{https://nick-jhlee.netlify.app/} }{jh\_lee00@kaist.ac.kr}{https://orcid.org/0000-0002-3898-6464}{}
\author{Laura Schmid}{Kim Jaechul Graduate School of AI, KAIST, Seoul, Republic of Korea}{laura.schmid@kaist.ac.kr}{https://orcid.org/0000-0002-6978-7329}{}
\author{Se-Young Yun}{Kim Jaechul Graduate School of AI, KAIST, Seoul, Republic of Korea \and \url{https://fbsqkd.github.io/}}{yunseyoung@kaist.ac.kr}{https://orcid.org/0000-0001-6675-5113}{}
\authorrunning{J. Lee, L. Schmid, and S.-Y. Yun} 

\Copyright{Junghyun Lee, Laura Schmid, and Se-Young Yun} 



\category{} 

\supplement{}
\supplementdetails[subcategory={Source Code}]{Software}{https://github.com/nick-jhlee/bandits-network}

\funding{JH and SY were supported by the Institute of Information \& Communications Technology Planning \& Evaluation (IITP) grants funded by the Korean government(MSIT) (No.2022-0-00311, Development of Goal-Oriented Reinforcement Learning Techniques for Contact-Rich Robotic Manipulation of Everyday Objects; No.2019-0-00075, Artificial Intelligence Graduate School Program(KAIST)). LS was supported by the Stochastic Analysis and Application Research Center (SAARC) under National Research Foundation of Korea grant NRF-2019R1A5A1028324.}

\acknowledgements{The authors thank Ulrich Schmid (TU Wien) and the anonymous reviewers for helpful comments and suggestions.}

\nolinenumbers 

\EventEditors{Alysson Bessani, Xavier D\'efago, Yukiko Yamauchi, and Junya Nakamura}
\EventNoEds{4}
\EventLongTitle{27th International Conference on Principles of Distributed Systems (OPODIS 2023)}
\EventShortTitle{OPODIS 2023}
\EventAcronym{OPODIS}
\EventYear{2023}
\EventDate{December 6--8, 2023}
\EventLocation{Tokyo, Japan}
\EventLogo{}
\SeriesVolume{286}
\ArticleNo{20}

\begin{document}

\maketitle

\begin{abstract}
Multi-armed bandits are extensively used to model sequential decision-making, making them ubiquitous in many real-life applications such as online recommender systems and wireless networking. We consider a multi-agent setting where each agent solves their own bandit instance endowed with a different set of arms. Their goal is to minimize their group regret while collaborating via some communication protocol over a given network. Previous literature on this problem only considered arm heterogeneity and networked agents separately. In this work, we introduce a setting that encompasses both features. For this novel setting, we first provide a rigorous regret analysis for a standard flooding protocol combined with the classic UCB policy. Then, to mitigate the issue of high communication costs incurred by flooding in complex networks, we propose a new protocol called Flooding with Absorption (FwA). We provide a theoretical analysis of the resulting regret bound and discuss the advantages of using FwA over flooding. Lastly, we experimentally verify on various scenarios, including dynamic networks, that FwA leads to significantly lower communication costs despite minimal regret performance loss compared to other network protocols.
\end{abstract}

\section{Introduction}
\label{sec:introduction}
Exploration-exploitation dilemmas form the basis of many real-life decision-making tasks~\cite{march1991exploration,levinthal1993myopia}.
In fact, the trade-off between making a choice to either stay with a current action or explore new possibilities appears as a feature in a variety of well-known applications~\cite{berger2014exploit,mehlhorn2015unpacking}. 
As a result, the {\it multi-armed bandit (MAB)} problem, which is designed to reflect dilemmas of this kind, has been intensely studied in a wide range of scenarios~\cite{bubeck-survey,lairobbins85,lattimore2020bandit}.
In the baseline setting, an agent must make sequential decisions by choosing from a set of possible actions (the ``arms'' of the bandit). In the setting of stochastic MABs, each arm gives rewards following an unknown probability distribution~\cite{bubeck-survey}.
Here, the goal is to minimize the cumulative \emph{regret} over some timespan $T$, i.e., the difference between the accumulated reward and the reward that arises from choosing only the best arm.
To reach this goal, the agent must balance exploring new actions and choosing already tested ones~\cite{Auer02,lairobbins85,bubeck-survey}.

For applications involving large-scale decentralized decision-making~\cite{landgren2021distributed}, such as online advertising, search/recommender systems, and wireless channel allocation, collaborative multi-agent multi-armed bandits are a natural modeling choice~\cite{anandkumar2010spectrum,sugawara2004foraging,li2010contextual,li2014cooperative,cesa2020cooperative,avner2016comm,jin2018robots,xia2020fed,li2020wireless}. In this setting, each agent plays their own bandit instance and communicates some information to others to minimize the {\it group} regret.
One common assumption in the literature is that agents share the same set of arms~\cite{kolla2018collaborative,dubey2020coorperative,wang2020collision}. However, arm homogeneity does not hold in many large-scale systems (e.g., contextual recommender systems), where agents often have heterogeneous arm sets of available actions~\cite{yang2022heterogeneous,chawla2023heterogeneous}.
For instance, in a distributed recommender system scenario, arms might correspond to the contents shown to users, such as movies, and rewards to user opinions; then, depending on one's location, the set of available movies for each agent may be different due to external constraints such as copyright issues.
In this case, it would be desirable for the service provider if all the individual systems with partially overlapping contents collaborate with one another to minimize the group regret.

Another common assumption is that agents are connected by a complete network, where agents can directly communicate with every other agent~\cite{buccapatnam2015info,yang2022heterogeneous}.
However, in real large-scale systems, 
agents are usually connected via a multi-hop communication network, where only adjacent nodes can exchange messages. To disseminate information to agents at larger distances here, agents need to forward messages. This is typically done by means of a flooding protocol~\cite{dubey2020coorperative,madhushani2021fault,kolla2018collaborative,wang2020collision} or gossiping protocol~\cite{szorenyi2013gossip,sankarasync,chawla2020gossiping,vial2021robust,chawla2023heterogeneous}; that is, a received message is forwarded to all neighbors or only one randomly selected neighbor, respectively.
Such a modeling assumption is ubiquitous in many real-life scenarios involving mobile/vehicular networks or social networks, among others.


\medskip
\noindent \emph{Contributions.}
To the best of our knowledge, the setting of collaborative, {\it heterogeneous} multi-agent multi-armed bandits communicating over a general {\it network} has not been investigated yet. Yet, this setting arises in a wide range of real-life applications, e.g. wireless channel allocation, where not all nodes on an underlying network can access the same channels. For such scenarios, one can neither assume fully connected or particularly regular communication topologies, nor homogeneous arm sets.
Distinct from previous work on multi-agent bandits, we here aim at efficient {\it network protocol design} for this novel setting; specifically, we want to design a {\it simple} alternative to flooding that achieves low communication complexity while retaining minimal loss in regret. Note that our research objective forces us to consider the bandit instances and the communication protocol in an integrated fashion, which is in stark contrast to the approaches used both in the existing bandit and networking literature, and thus contributes to the novelty of our approach. 

To address the significant issue of exploding communication complexity of flooding in our setting, we introduce a new lightweight communication protocol for complex networks, called \textsc{Flooding with Absorption (FwA)}. Its design principle is inherently coupled with the given bandit instances.
We provide theoretical and experimental results showing that this protocol is highly communication-efficient on a wide range of complex networks, yet induces minimal regret performance loss for complex network topologies, even compared to standard flooding.
Using \textsc{FwA} can also help avoid heavy individual link congestion in complex networks.
An important practical advantage of our protocol is that it is fully agnostic to the network structure, and can therefore be deployed on dynamically changing networks without any need for fine-tuning.



\begin{figure}[t]
    \centering
    \resizebox{0.9\linewidth}{!}{
        \includegraphics{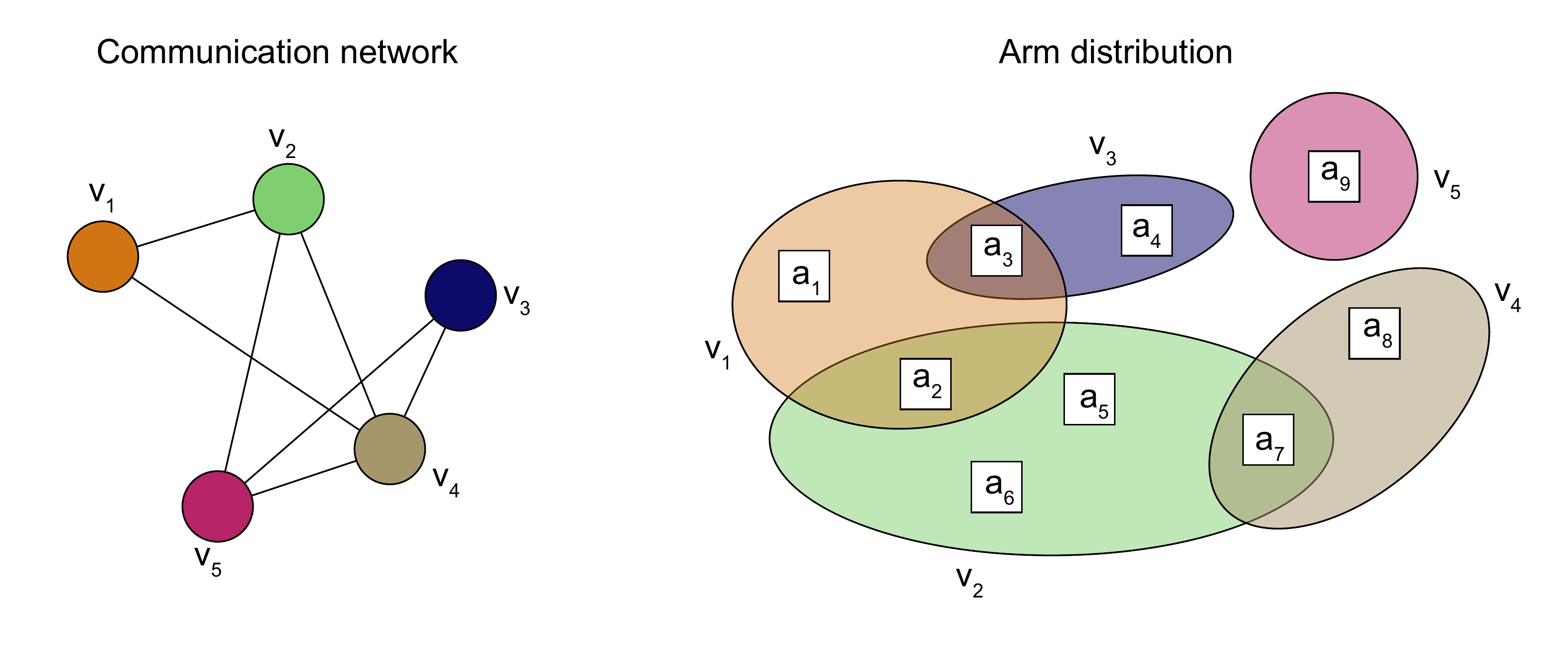}
    }
    \caption{Communication network and arm heterogeneity.
    }
    \label{fig:system-model}
\end{figure}

\section{System Model}
\label{sec:system-model}

We now describe our setting of collaborative\footnote{We remark that we do {\it not} consider any collisions \cite{liu2010collision,lelarge2013collision,wang2020collision} where two neighbors pulling the same arm do not affect their observed rewards in any way. Rather, we focus on the collaborative setting where the agents are encouraged to cooperate with one another by sharing their own observations.} \emph{heterogeneous} multi-agent multi-armed stochastic bandits over a general communication network; see Figure~\ref{fig:system-model} for an illustration.
We assume that there are $N$ agents connected by an undirected graph $\gG = (\gV, \gE)$, with $|\gV| = N$.
We denote by $\gN_\gG(v)$ be the neighborhood of $v$ in $\gG$ {\it not} including $v$, and by $\gG[S]$ the induced subgraph for $S \subset \gV$.
Also, for an integer $\gamma \geq 1$, the $\gamma$-th order graph power of $\gG$, denoted as $\gG^\gamma = (\gV, \gE^\gamma)$, is defined as the graph on $\gV$ such that $\{v, w\} \in \gE^\gamma$ iff $d_\gG(v, w) \leq \gamma$, where $d_\gG(v, w)$ is the length of the shortest path in $\gG$ connecting $v$ and $w$.
Each agent $v \in \gV$ has access to a finite set $\gK_v$ of arms with cardinality $K_v$ that they can pull; let $\gK = \cup_{v \in V} \gK_v$ be the total set of arms of cardinality $K$.

Following \cite{lattimore2020bandit}, let $\gM_\sigma$ be a set of $\sigma$-sub-Gaussian distributions, and let $\mu : \gM_\sigma \rightarrow \sR$ be a function mapping each (reward) distribution to its mean.
Each arm $a \in \gK$ is associated with an unknown reward distribution $P_a \in \gM_\sigma$.
For simplicity, let $\mu_a := \mu(P_a)$.
We note that $P_a$ is independent of the agents' identities, i.e., each agent $v$, regardless of their arm set $\gK_v$, faces the same distribution of rewards for the same arm $a$ (whenever $\gK_v$ contains $a$), and receives an i.i.d. reward from this distribution upon pulling this arm. 
We denote by $a_\star^v$ the best {\it local} arm for agent $v$ that satisfies $\mu_{a_\star^v} > \mu_a$ for all $a \in \gK_v \setminus \{a_\star^v\}$, and set $\mu_\star^v = \mu_{a_\star^v}$.
The main challenge in the regret analysis is that even for the same arm $a$, the suboptimality gap may be different across agents containing $a$.

The execution of all agents proceeds in a sequence of synchronous rounds $t=1,2,\ldots$.
In each round $t$, all agents simultaneously (i) pull some arm, (ii) compute and send a message to their neighbors, and (iii) receive and process all messages from their neighbors.
From the perspective of agents, let us denote by $\gV_a = \{ v \in \gV : a \in \gK_v \} \subseteq \gV$ the set of agents having action $a$, and let $\gV_{-a} \subseteq \gV_a$ be the set of agents containing $a$ as a {\it suboptimal} arm, i.e., $a \neq a_\star^v$.

As done in the classic work on regret minimization in collaborative multi-agent bandits \cite{kolla2018collaborative,yang2022heterogeneous,madhushani2021fault}, our goal is to minimize the expected {\it group} regret at time horizon $T$, $\E[R(T)]$, defined as:
\begin{equation}
    \E[R(T)] := \sum_{v \in \gV} \E[R^v(T)], \quad \E[R^v(T)] := \sum_{a \in \gK_v} \Delta_a^v \E[N_a^v(T)],
\end{equation}
where $\Delta_a^v := \mu_\star^v - \mu_a$ is the agent-specific gap of arm $a$ and $N_a^v(t)$ is the number of times agent $v$ plays the arm $a$ up to time $t$.

We note that due to the two sources of heterogeneity in our setting, the benefit one receives from collaboration differs; 
furthermore, the same arm $a$ may be optimal for some agents but suboptimal for others.
Of course, even so, we expect that agents communicating and collaborating should lead to a speed-up (in terms of the group regret) compared to the baseline without any information sharing.
The question is then how much speed-up one could get from collaboration under our heterogeneous system model. An additional prominent issue for realistic applications and, in particular, \emph{complex} networks concerns the communication complexity of the information-sharing protocol used. Designing communication protocols with {\it low} communication complexity, defined as the number of messages sent and forwarded, is of paramount importance to not overshadow the improvement in regret due to collaboration.
\section{Flooding}
As is common in much of the previous literature, we will focus on agents that individually run the classic upper-confidence bound (UCB) policy~\cite{Auer02,kolla2018collaborative,yang2022heterogeneous} under which each agent pulls the arm associated with the maximum of the so-called UCB index, which is the sum of empirical reward (up to time $t$) and an additional bonus term.
Agents can, and should, take advantage of the distributed setting by communicating pulled arms and received rewards amongst each other over the underlying communication network.
This is not straightforward to implement or analyze, given that the agents' arm sets are all different, and thus, even when broadcasting over multiple hops by flooding the network is available, the speed-up gained from this is not immediate.
In this section, we review (and reformulate) the standard flooding protocol for our setting and discuss its properties.

\SetKwFunction{UCB}{UCB}
\SetKwFunction{Receive}{Receive}
\SetKwFunction{Store}{Store}
\SetKwProg{Fn}{Function}{:}{}

\noindent
\vspace{0pt}
\begin{figure*}[t]
\begin{minipage}[t]{0.6\textwidth}
\begin{algorithm}[H]
{\footnotesize
\label{alg:ucb-flooding}
\caption{\textsc{Flooding, \textsc{FwA} with UCB}}
\KwInput{$\gamma \in \sN$, $\alpha > 0$, ${\color{blue}absorb} \in \{True, False\}$, $f : \sN \rightarrow [1, \infty)$}
Initialize $\gM^v = queue()$, $\gH^v = queue(maxlen=\gamma N)$\;
$M^v_a(0) = 0, \hat{\mu}_a^v = 0$ for all $(v, a) \in \gV \times \gK$ \;
\For{$t \in [T]$}{
    \For{$v \in \gV$}{
        \eIf{$t \leq K_v$}{
            play the $t$-th arm of $\gK_v$\;
        }{
            play the arm $a^v(t)$ chosen as follows:
            $\displaystyle
    a^v(t) = \argmax_{a \in \gK_v}\left\{ \hat{\mu}_{a}^v(t) + \sqrt{\frac{ 2 \log f(t) }{M^v_a(t)}}\right\}
    $\;
        }
        $hash \gets \textsc{Hash}(v, a^v(t), X^v(t))$\;
        $\gM^v.push(\langle a^v(t), X^v(t), hash, v, \gamma \rangle)$\;
    }
    \For{$v \in \gV$}{
        \While{$\widetilde{\gM}^v$ is not empty}{
            $m \gets \widetilde{\gM}^v.pop()$\;
            \For{$w \in \gN_\gG(v) \setminus \{m(3)\}$}{
                $m' \gets copy(m)$\;
                \If{$m'(2) \not\in \gH^w$}{
                    $\gH^w.push(m'(2))$\;
                    \Receive($w$, $m'$, ${\color{blue}absorb}$)\;
                }
                delete $m'$
            }
            delete $m$\;
        }
    }
}
}
\end{algorithm}
\end{minipage}
~~~
\begin{minipage}[t]{0.525\textwidth}
\begin{algorithm}[H]
{\footnotesize
\caption{Subroutines}\label{alg:subroutines}
\Fn{\Receive{$w$, $m$, ${\color{blue}absorb}$}}{
    \eIf{$m(0) \in \gK_w$}{
        $M^w_{m(0)}(t) \gets M^w_{m(0)}(t - 1) + 1$, 
        $\hat{\mu}_{m(0)}^w(t) \gets \frac{M^w_{m(0)}(t-1) \hat{\mu}_{m(0)}^w(t-1) + m(1)}{M^w_{m(0)}(t)}$\;
        \eIf{${\color{blue}absorb}$ is True}{
            delete $m$\;
        }{
            \Store($w$, $m$)\;
        }
    }{
        \Store($w$, $m$)\;
    }
}
\Fn{\Store{$w$, $m$}}{
    $m(4) \gets m(4) - 1$\;
    \eIf{$m(4) < 0$}{
        delete $m$\;
    }{
        $m(3) \gets w$\;
        $\gM^w.push(m)$\;
    }
}
}
\end{algorithm}
\end{minipage}
\end{figure*}

\subsection{Flooding}
\label{sec:flooding}
\textsc{Flooding} (broadcasting) lets each agent send all messages to all its neighbors in every round, with the number of times the message is forwarded limited by the \emph{time-to-live} (TTL) $\gamma$~\cite{chang2007controlled,tanenbaum2003computer,vojnovic2011hop,rahman2004controlled}.
To account for potential loops in the network and avoid a broadcast storm~\cite{tseng2002storm,wisitpongphan2007storm}, we explicitly use a sequence number-controlled flooding (SNCF) variant. 
The pseudocode of \textsc{Flooding} along with each agent's UCB algorithms is presented in Algorithm \ref{alg:ucb-flooding} with ${\color{blue}absorb} = False$.

\textsc{Flooding} proceeds as follows: In each round $t$, each agent $v$ pulls an arm $a^v(t)$ that has the highest upper confidence bound (line 8).
Note that $M^v_a(t)$ is the number of pulls of arm $a$ available to agent $v$ by time $t$, and $\hat{\mu}_{a}^v$ is the estimate of $\mu_a$ made by agent $v$ at time $t$.
In both estimates, agent $v$ uses all observations available to $v$ by time $t$, including the messages relayed to them.
Having received the corresponding reward $X^v_{a^v(t)}(t)$ from pulling arm $a^v(t)$, agent $v$ creates a message
\begin{equation}
   m=\langle a^v(t), X^v_{a^v}(t), \textsc{Hash}(v, a^v(t), X_{a^v}^v(t)), v, \gamma \rangle,
\end{equation}
and pushes it to $\gM^v$, its current queue of messages to be sent (line 11).
After UCB has been completed, each agent sends and receives messages to and from its neighbors (lines 14-21).

Our message $m = \langle m(0), m(1), m(2), m(3), m(4) \rangle$ consists of the following components:
$m(0)$ and $m(1)$ are the arm pulled by agent $v$ and the reward received at time $t$, respectively.
$m(2)$ is a hash value of the originating agent $v$, the arm pulled, and the obtained reward that acts as a unique identifier of the message.
Our protocol uses $m(2)$ to control flooding by avoiding routing loops that can lead to broadcast storms and improper bias in the reward estimations (the estimation protocol is shown in line 3 of Algorithm~\ref{alg:subroutines}).
Each agent $v$ keeps track of the hash values of messages that they have seen until time $t$ via a {\it queue}\footnote{The queue operations used in the algorithm ($pop()$, $push()$, $flush()$) are defined as usual~\cite{Cormen22}.} of size $\gamma N$, denoted as $\gH^v$.
If an already-seen message comes in (line 18), that message is deleted on arrival (line 22).
The memory length of $\gamma N$ is the worst-case space complexity that arises from keeping track of all messages from all agents for the last $\gamma$ time steps, as all messages can be forwarded at most $\gamma$ times.
$m(3)$ is the agent that last forwarded the message; if the receiver $w$ 
passes on $m$, they replace $m(3)$ with $w$ and forward $m$ to all neighbors except the originator $m(3)$ (line 16).
This prevents messages from echoing after one hop.
$m(4)$ keeps track of the remaining life span of the message (TTL), which is initialized to $\gamma$.
It is decayed by $1$ every time a message is forwarded, and the message is discarded when TTL reaches $0$. 
We note that $\gamma=1$ is equivalent to \textsc{Instant Reward Sharing} (\textsc{IRS})~\cite{kolla2018collaborative}, where each agent only sends its message to its neighbors, and any message containing arm $a$ that is sent to agents not containing $a$ becomes void.

We assume that the nodes have no knowledge of the network topology, or who their neighbors are, which is a realistic assumption in complex networks, and wireless networks.
This is in contrast to some of the previous works, e.g., \cite{madhushani2021fault}, which assumes that each agent knows its neighborhood in $\gG^\gamma$, which essentially bypasses any issues regarding communication complexity. Specifically, this also abstracts away any difficulties arising from delayed messages traveling along different paths.


\subsection{Group Regret Analyses of \textsc{Flooding}} 
The regret bound of stochastic bandits generally depends on a problem-dependent quantity~\cite{Auer02} that quantifies the difficulty of learning. For instance, for a single-agent multi-armed stochastic bandit, the regret bound scales as $\sum_{a \in \gK \setminus \{a^\star\}} \frac{\log T}{\Delta_a}$, where $\Delta_a$ is the gap between the mean rewards of best arm $a^\star$ and the suboptimal arm $a$.
The intuition is that if the mean rewards are similar, one would need a much tighter confidence interval to identify the optimal arm, forcing one to pull $a$ many more times, precisely inversely proportional to its reward gap.
This is known to be asymptotically optimal~\cite{lairobbins85}.

In our setting, we would expect our problem-dependent quantity to depend on both the underlying network topology and arm distribution. To see this, given a graph $\gG = (\gV, \gE)$, we first recall some graph-theoretic quantities~\cite{Bollobas02}:
\begin{definition}
    The {\bf clique covering number} of $\gG$, denoted as $\theta(\gG)$, is the smallest size of a partition of $\gV$ such that each part induces a clique.
    Any such partition (not necessarily minimum) is called a {\bf clique cover}.
    The {\bf independence number} of $\gG$, denoted as $\alpha(\gG)$, is the maximum size of a subset of $\gV$ that induces no edges.
\end{definition}

We now define our problem-dependent quantity $\Delta_a^\gamma$ as follows:
\begin{equation}
\label{eqn:delta}
    \Delta_a^\gamma := \max_{\gC_{-a}} \left( \sum_{C \in \gC_{-a}} \left\{ \frac{2}{\min_{v \in C} \Delta_a^v} - \frac{1}{\max_{v \in C} \Delta_a^v} \right\} \right)^{-1},
\end{equation}
where $\Delta_a^v = \mu_\star^v - \mu_a$ is the agent-specific suboptimality gap of arm $a$, and $\max_{\gC_{-a}}$ is taken over all possible clique covers $\gC_{-a}$ of $[\gG^\gamma]_{-a} := \gG^\gamma[\gV_{-a}]$.
If $\gV_{-a} = \emptyset$, we set $\Delta_a^\gamma = 0$.

We now present the nonasymptotic regret upper bound for \textsc{Flooding}:
\begin{theorem}
\label{thm:ucb1}
    Algorithm \ref{alg:ucb-flooding} with ${\color{blue}absorb}=False$, $f(t) = t^\alpha$, $\alpha > \max\left( \frac{1}{2}, \frac{2\sigma^2}{\gamma + 1} \right)$, and $\gamma \in \{1, \cdots, \diam(\gG)\}$ achieves the group regret upper bound
    \begin{equation}
    \label{eqn:regret1}
        \E\left[ R(T) \right] \leq \sum_{\substack{a \in \gK \\ \Delta_a^\gamma > 0}} \frac{4\alpha \log T}{\Delta_a^\gamma} + b(\gamma) + \sum_{a \in \gK} f_a(\gamma),
    \end{equation}
    where
    \begin{equation*}
        b(\gamma) := \left( \frac{\alpha + 1/2}{\alpha - 1/2} \right)^2 \frac{8(\gamma + 1)}{\log\frac{(\gamma + 1) (\alpha + 1/2)}{4\sigma^2}} \sum_{a \in \gK} \sum_{v \in \gV_a} \Delta_a^v, \quad
        f_a(\gamma) = \sum_{v \in \gV_{-a}} \Delta_a^v \min\left( 2\gamma, \frac{4 \alpha \log T}{(\Delta_a^v)^2}\right).
    \end{equation*}
\end{theorem}
The complete proof, deferred to Appendix~\ref{proof:thm1}, uses a clique covering argument and an Abel transformation.
See Appendix~\ref{app:challenges} for a discussion of the main technical challenges when proving the regret bound for our setting, compared to previously considered settings.

By choosing $\gC_{-a}$ as the minimum clique cover for each $a$ in the definition of $\Delta_a^\gamma$, a simplified, asymptotic regret bound can be deduced:
\begin{corollary}
\label{cor:ucb1}
    When $\max\left(b_{\alpha, \gamma, \sigma}, \sum_{a \in \gK} f_a(\gamma) \right) = o(\log T)$,
    \begin{equation}
    \label{eqn:regret2}
        \limsup_{T \rightarrow \infty} \frac{\E\left[ R(T) \right]}{\log T} \leq \sum_{\substack{a \in \gK \\ \Delta_a^\gamma > 0}} \frac{4\alpha}{\Delta_a^\gamma}
        \leq \sum_{\substack{a \in \gK \\ \tilde{\Delta}_a > 0}} \frac{8\alpha \theta([\gG^\gamma]_{-a})}{\tilde{\Delta}_a},
    \end{equation}
    where $\tilde{\Delta}_a := \min_{v \in \gV_{-a}} \Delta_a^v$.
\end{corollary}
Note that 
%
$\tilde{\Delta}_a$ is the suboptimality gap introduced in Yang et al. \cite{yang2022heterogeneous}, where they studied the setting of heterogeneous bandits on a fully-connected network.
When $\gG$ is a fully connected network,  $\theta(\cdot) = 1$, and we recover their regret bound, with an improved constant in $\alpha$.

\medskip
Since our general setting also applies to the restricted cases considered in the previous literature, we can compare our regret bounds to existing ones.
In the same setting without collaboration, the group regret scales as $\sum_{a \in \gK} \frac{|\gV_{-a}| \log T}{\tilde{\Delta}_a}$; thus when $\tilde{\Delta}_a$ is considered to be constant, the regret always scales linearly in $N$.
Compared to this, depending on the network, the regret bound of \textsc{Flooding} scales with the clique covering number of subgraphs of $\gG^\gamma$, which is usually strictly less than $N$, and in some cases, even sublinear.

Similarly, our regret bounds and our problem-dependent difficulty quantity $\Delta_a^\gamma$ also generalize previous literature on collaborative multi-agent multi-armed bandits.
When the network is a clique, we recover the regret bound presented in \cite{yang2022heterogeneous} with matching $\log T$ dependency and an improved leading coefficient\footnote{Theorem 2 of \cite{yang2022heterogeneous} requires $\alpha > 2$, and thus with proper scaling, it can be seen that our coefficient is $8\alpha$ while their coefficient is $24\alpha$.}.
In the homogeneous agents setting with a general network topology~\cite{kolla2018collaborative,madhushani2021fault}, $\Delta_a^\gamma$ reduces to $\frac{\Delta_a^{Kolla}}{\theta(\gG^\gamma)}$, where $\Delta_a^{Kolla}$ is the suboptimality gap as defined in \cite{kolla2018collaborative}, satisfying $\Delta_a^{Kolla} := \Delta_a^v$ for all $v \in \gV$.
As $\frac{1}{\theta(\gG^\gamma)}$ is independent of the arm $a$, we have shown that our $\Delta_a^\gamma$ successfully generalizes the suboptimality gap of~\cite{kolla2018collaborative}.
When $\gamma = \diam(\gG)$, we have that $\Delta_a^\gamma = \Delta_a^{Kolla}$ as $\theta(\gG^\gamma) = 1$, which results in the same regret bound as in~\cite{kolla2018collaborative}.

When $\gamma = 1$ (\textsc{IRS}), it can be observed that \textsc{IRS} and \textsc{FwA} coincide, yet our bound is a bit worse compared to \cite{kolla2018collaborative}, whose bound depends on $\alpha(\gG)$.
We believe that such a gap is an inherent artifact of our proof, and we leave closing this gap to future work. Also, we should remark that the difference between the two regret bounds for \textsc{IRS} is generally small, as the gap depends on the covering gap $\theta(\gG) - \alpha(\gG)$, which is known to be small for many classes of graphs, and zero for perfect graphs~\cite{gyarfas1987perfect}; see \cite{gyarfas2012gap,munaro2017gap} for some recent advances.

\subsection{Drawbacks of \textsc{Flooding}}
Flooding leads to optimal information dissemination, and thus significantly improves the group regret.
Yet, it is very expensive in terms of communication complexity, defined as the {\it cumulative} number of messages sent by all agents~\cite{yang2022heterogeneous}.
Indeed, for $\gamma = \diam(\gG)$, the worst-case communication complexity is $\gamma N |\gE| T$, which is attained when every message created by every agent up to time $T$ is being passed around at every edge. This can quickly congest the network.
One na\"{i}ve way of controlling the communication complexity is to set the TTL, $\gamma$, to a low enough value. 
However, in our setting, the trade-off between communication complexity and group regret is not trivial due to the arm heterogeneity; for instance, \textsc{IRS}~\cite{kolla2018collaborative,madhushani2021fault}, i.e., $\gamma = 1$, has a lower message complexity $N |\gE| T$ but often does not result in good regret guarantees, as immediate neighbors may not share any arms. On the other hand, (uniform) gossiping algorithms for bandit problems~\cite{chawla2020gossiping,sankarasync,shah2009gossip} 
suffer from large latencies on networks with sparse links~\cite{haeupler2015gossip,censor2012rumor}. 

It is thus desirable to have a simple communication protocol with good regret guarantees when combined with the UCB policy (compared to \textsc{Flooding}) and low communication complexity. With this, we can target more complex network structures commonly found in real-life applications.
For such settings, we introduce a new protocol that interpolates between the communication-efficient nature of \textsc{IRS} and the regret of \textsc{Flooding} by using the intrinsic heterogeneity of the system caused by network topology and arm distributions.

\section{A New Efficient \& Effective Protocol on Complex Networks: Flooding with Absorption}
\label{sec:FwA}
\begin{figure}
    \centering
     \resizebox{1.05\linewidth}{!}{
     \includegraphics{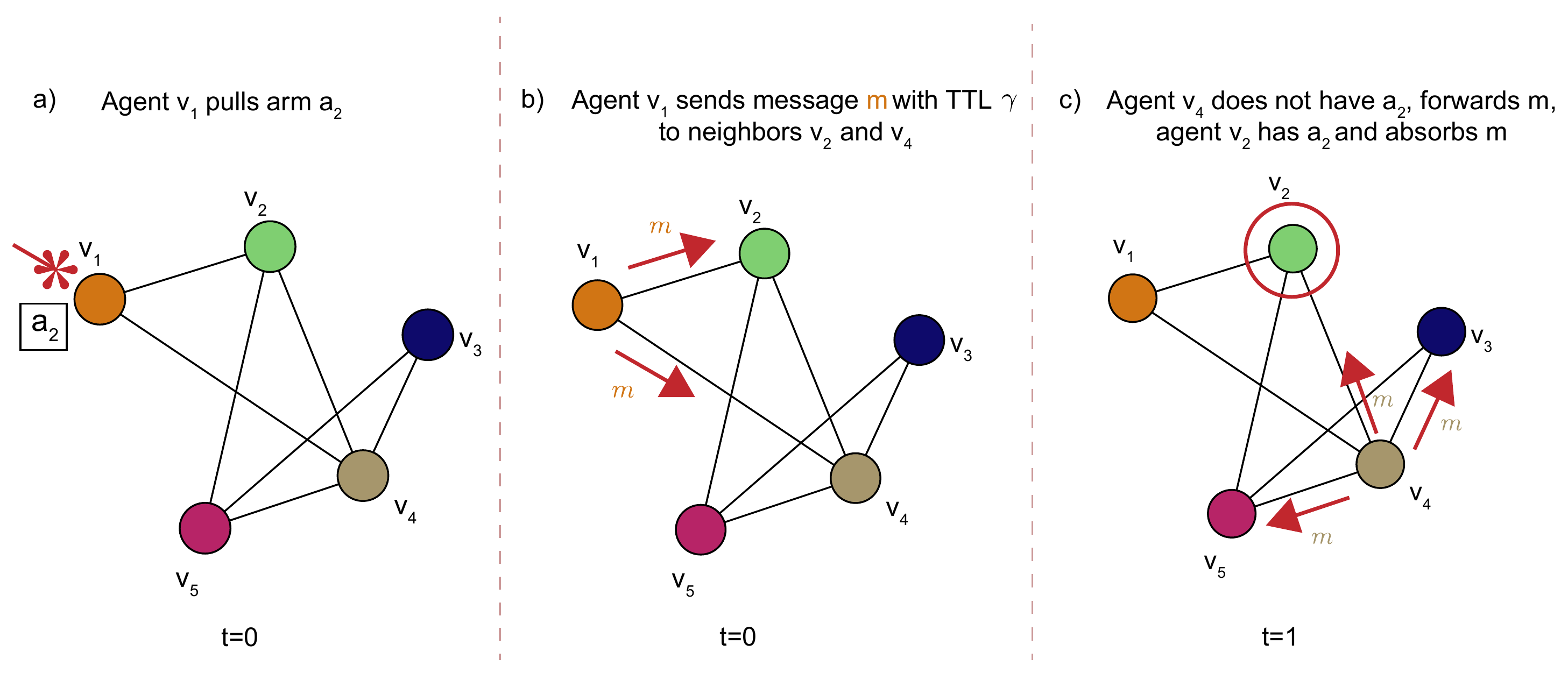}
    }
    \caption{{\bf Flooding with Absorption (FwA)}.
    {\bf a, }An agent ($v_1$) pulls one of its arms ($a_2$).
    {\bf b, }$v_1$ sends a message $m$ to its neighbors, with a TTL $\gamma$.
    {\bf c, }Since one receiver of the message ($v_4$) does not have $a_2$ in its arm set, they forward $m$ to their neighbors except the originator $v_1$. The other receiver ($v_2$) has arm $a_2$ in their arm set, and thus it absorbs $m$.}
    \label{fig:hmp}
\end{figure}

In this section, we propose a new approach, which we call {\bf \textsc{Flooding with Absorption (FwA)}} (Figure~\ref{fig:hmp}), whose pseudocode is shown in Algorithm \ref{alg:ucb-flooding} with ${\color{blue} absorb} = True$. 
In contrast to \textsc{Flooding}, once a message (some copy of it to be precise) containing arm $a$ reaches an agent whose arm set includes the arm $a$, the agent {\it absorbs} that message, i.e., does not forward it any further.
Additionally, as in \textsc{Flooding}, we retain the TTL $\gamma$, meaning that if a message originating at time $t$ has not found an absorbing agent until $t'=t+\gamma$, it gets discarded.
It is also dropped if the message hits a ``dead end,'' i.e., a leaf node.

This seemingly small difference to \textsc{Flooding} is critical in ensuring low communication complexity, as it prevents messages from circulating for too long.
We note that \textsc{FwA} is somewhat reminiscent of the well-studied replication-based epidemic- and other controlled flooding algorithms~\cite{vahdat2000epidemic,rahman2004controlled,lim2001flooding,eugster2004epidemic}, which were designed for various networking applications, e.g., ad-hoc networks.
Our \textsc{FwA} protocol distinguishes itself by {\it using} the inherent heterogeneity of agents {\it without} any explicit tuning or need for solving NP-hard combinatorial problems~\cite{lim2001flooding}.
Furthermore, the goal of \textsc{FwA} is to disseminate information, which is generated at each timestep, to nodes that may benefit from it for its learning, {\it not} to route packets from an arbitrary source to an arbitrary destination. Despite some outward protocol similarity, it hence also differs from classic P2P systems such as Gnutella~\cite{gnutella02,lv2022gnutella}, where there is no such intrinsic correlation between sender and receiver. In the bandit setting, \textsc{FwA} is advantageous because the sender and final receiver share the arm in question.


\begin{remark}
    Each agent must have a sufficiently large memory buffer to store the messages to be sent in the next round and previously seen message identifiers.
    As all messages expire after $\gamma$ rounds, this memory requirement is at most $\gamma N$.
    Also, we note that the communication complexity of \textsc{FwA} ranges from $N|\gE|T$ to $\gamma N |\gE| T$, depending on the underlying network topology and the arm distribution.
\end{remark}

\subsection{Group Regret Bound of \textsc{FwA}}
\label{sec:fwabound}
As \textsc{FwA} is algorithmically similar to \textsc{Flooding}, their regret bounds are also somewhat similar.
To formalize this, we first consider a graph $\gG = (\gV, \gE)$, and let $\vc : \gV \rightarrow 2^\gK$ be a multi-coloring with overlap allowed, i.e., it may be that $\vc(v) \cap \vc(w) \not= \emptyset$ for $\{v, w\} \in \gE$.
Let $a \in \gK$ and $v, w \in \gV$ be such that $a \in \vc(v) \cap \vc(w)$.
\begin{definition}
    A path $v_0 v_1 \cdots v_m$ (of length $m$) is said to {\bf $a$-free} if $a \not\in \bigcup_{i \in [m-1]} \vc(v_i)$, where $[m]=\{0,1,...m\}.$
\end{definition}
\begin{definition}
\label{def:nonblocking}
    For $\gamma \geq 1$ and $a \in \gK$, we define the {\bf $(a, \vc)$-non-blocking graph power of $\gamma$-th order of $\gG$}, denoted as $\gG^\gamma_{(a, \vc)}$, as a graph on $\gV$ with the edge set $\gE^\gamma_{(a, \vc)}$ such that $\{v, w\} \in \gE^\gamma_{(a, \vc)}$ iff there exists an $a$-free path from $v$ to $w$ in $\gG$ of length at most $\gamma$.
\end{definition}

\begin{remark}
    Def.~\ref{def:nonblocking} is similar to {\it color-avoiding percolation (CAP)} in statistical physics~\cite{krause2016color,krause2017color,kadovic2018color}, albeit there are several differences.
    We consider a multi-coloring $\vc : v \mapsto \gK_v$, while CAP is only studied for a single color per vertex.
    Also, CAP only considered the criticality of the connectivity, while in our case, the performance of our algorithm depends on specific graph invariants (e.g., chromatic number) of color-dependent subgraphs.
\end{remark}

Defining the suboptimality gap $\Delta_a^{FwA}$ for \textsc{FwA} as
\begin{equation}
\label{eqn:delta2}
    \Delta_a^{FwA,\gamma} := \max_{\gC_{(a, \vc)}^\gamma} \left( \sum_{C \in \gC_{(a, \vc)}^\gamma} \left\{ \frac{2}{\min_{v \in C} \Delta_a^v} - \frac{1}{\max_{v \in C} \Delta_a^v} \right\} \right)^{-1},
\end{equation}
where $\min_{\gC_{(a, \vc)}^\gamma}$ is over all possible clique covers $\gC_{(a, \vc)}^\gamma$ of $\gG_{(a, \vc)}^\gamma$, it is easy to see that the following theorem holds:
\begin{theorem}
\label{thm:ucb2}
    With ${\color{blue} absorb}=True$, Theorem \ref{thm:ucb1} holds with $\Delta_a^\gamma$ replaced by $\Delta_a^{FwA,\gamma}$.
\end{theorem}
Similarly, with an appropriate choice of the clique cover, we have the following simplified asymptotic regret bound:
\begin{corollary}
\label{cor:ucb2}
    With the same assumption as in Theorem~\ref{thm:ucb1}, we have 
    \begin{equation}
    \label{eqn:regret4}
        \limsup_{T \rightarrow \infty} \frac{\E\left[ R(T) \right]}{\log T} \leq \sum_{\substack{a \in \gK \\ \Delta_a^\gamma > 0}} \frac{4\alpha}{\Delta_a^{FwA,\gamma}}
        \leq \sum_{\substack{a \in \gK \\ \tilde{\Delta}_a > 0}} \frac{8\alpha \theta\left([\gG_{(a, \vc)}^\gamma]_{-a}\right)}{\tilde{\Delta}_a},
    \end{equation}
    where $\tilde{\Delta}_a = \min_{v \in \gV_{-a}} \Delta_a^v$.
\end{corollary}

As $\gG^\gamma_{(a, \vc)}$ is always a subgraph of $\gG^\gamma$, it can be easily seen that the regret upper-bound of \textsc{Flooding} is always better than that of \textsc{FwA}.
But as we will demonstrate later, at the price of {\it slightly} worse regret, \textsc{FwA} obtains significantly better communication complexity than \textsc{Flooding}.
Corollary \ref{cor:ucb1} and \ref{cor:ucb2} imply that the gap in the asymptotic regret upper-bounds of \textsc{Flooding} and \textsc{FwA} roughly scales with $\sum_{\substack{a \in \gK \\ \tilde{\Delta}_a > 0}} \frac{\delta_a}{\tilde{\Delta}_a}$, where $\delta_a := \theta\left([\gG_{(a, \vc)}^\gamma]_{-a}\right) - \theta\left([\gG^\gamma]_{-a}\right)$.


To get some intuition, we consider two extreme cases.
First, suppose that the arms are so heterogeneous that no agents of distance at most $\gamma$ share arm $a$.
In this case, we have $\gG^\gamma_{(a, \vc)} = \gG^\gamma$, and $\delta_a = 0$.
Now suppose that all agents have the same arm set, i.e., $\gK_v = \gK$ for all $v \in \gV$, in which case \textsc{FwA} is equivalent to \textsc{IRS}, i.e., messages do not get forwarded beyond direct neighbors.
Hence, we have that $\gG^\gamma_{(a, \vc)} = \gG$ for all $\gamma \geq 1$ and $a \in \gK$, i.e., $\delta_a = \theta([\gG]_{-a}) - \theta([\gG^\gamma]_{-a})$.
Thus, for small $\gamma$'s and $\gG$ with {\it large} average path length~\cite{barabasi2002network} between agents containing $a$, $\delta_a$ is small.

\subsection{Advantages of Flooding with Absorption}
\label{sec:advantages}

We now informally argue the advantages of using \textsc{Flooding with Absorption} over other protocols such as \textsc{Flooding} or \textsc{IRS} when we run it on complex network topologies.

\medskip

\noindent \emph{Interpolation between IRS and Flooding.}
\textsc{FwA} naturally interpolates between \textsc{IRS} and \textsc{Flooding} in terms of information propagation, which is advantageous on complex networks that are not particularly regular, e.g., those with both dense and sparse regions.
In dense parts of the network, where many nodes share arms, \textsc{FwA} is closer to IRS: a message containing the shared arm and its reward gets absorbed quickly.
On the other hand, in regions of the network where the arm that a particular node pulled is rare, \textsc{FwA} acts like \textsc{Flooding} with $\gamma \gg 1$, thereby ensuring that agents at a larger distance get information that is relevant.
We additionally note that setting the TTL to larger values in \textsc{FwA} will always be less costly than doing so in \textsc{Flooding}, as the probability of congestion is much smaller.\\
\noindent \emph{Comparable Regret Guarantees.} As \textsc{FwA} acts as a mix of \textsc{IRS} and \textsc{Flooding}, its regret should be bounded by \textsc{IRS} (where messages get absorbed in just one step) from above, and \textsc{Flooding} (where messages are not absorbed until the TTL expires) from below.
The combination of Theorems \ref{thm:ucb1} and~\ref{thm:ucb2} gives us an expression for the gap between the regret upper bounds of ~\textsc{Flooding} and ~\textsc{FwA}.
From this, we can conclude that for the regret gap between \textsc{FwA} and \textsc{Flooding} to be small, either the graph is so sparse that the average path length~\cite{barabasi2002network} between agents {\it containing} $a$ is large, or the graph is dense but the arm distribution is sparse enough such that the same property holds.
We emphasize that although the gap may be nonzero, the exploding communication complexity of \textsc{Flooding} demonstrates a clear trade-off between performance and communication complexity.
On the other hand, it is also expected that \textsc{FwA} will outperform (uniform) gossiping in terms of the regret. 
In fact, in networks with sparse links connecting very dense network regions, the probability that a gossiping protocol hits the sparse link before the TTL expires can be arbitrarily small.\\
\noindent \emph{Communication Efficiency.}
Having messages absorbed by agents that can profit from its information implies that the \textsc{FwA} protocol completely falls back to the baseline \textsc{Flooding} algorithm only in the case of a network where particular arms are very rare.
This means that if there is a ball of radius $\gamma$ in the network in which two agents share an arm, the communication complexity of \textsc{FwA} will already be lower than that of \textsc{Flooding}, $O(N \cdot |\gE| \cdot \gamma \cdot T)$.
In networks of high density and few arms, the communication complexity of \textsc{FwA} will be close to that of \textsc{IRS}, i.e., lowered by a factor $\gamma$. 
Moreover, due to the arm-dependent absorptions, we expect that \textsc{FwA} will result in much lesser number of messages sent across each individual link per round.
Hence, \textsc{FwA} has the advantage of being able to mitigate network overload and heavy link congestion without much overhead or the need to resort to gossiping protocols, which is particularly salient for applications such as large-scale and wireless networks.\\
\noindent\emph{No tuning requirements.}
\textsc{FwA} also has an important practical advantage: it has no tunable parameters beyond the TTL $\gamma$, which means it is close to network agnostic.
This is in contrast to protocols like probabilistic flooding~\cite{oikonomou2023prob}, which stops message propagation with some constant probability $q$.
While such protocols can also reduce communication complexity to a scalable degree, the ``optimal'' stopping probability $q$ is highly instance-dependent, making them quite hard to use in unknown, real-life networks, in particular, dynamically changing networks.
\textsc{FwA}, on the other hand, can effectively deal with such instances, which we verify in Section~\ref{sec:experiments}.
\section{Regret Lower Bound}
\label{sec:lower-bound}
We consider a decentralized\footnote{see Appendix A of \cite{dubey2020coorperative} for the precise measure-theoretic definition of decentralized policies.} policy $\Pi = (\pi^v)_{v \in \gV}$, where $\pi^v : [T] \rightarrow \gP(\gK)$ is the agent-wise policy followed by agent $v$, possibly affected by other policies and the history.
For the regret lower bound, we consider a rather general class of policies satisfying the following property, which has been widely adapted in bandit literature~\cite{lairobbins85,kolla2018collaborative,dubey2020coorperative}:
\begin{definition}
    $\Pi$ is said to be {\bf individually consistent} if, for any agent $v$ and any $a \in \gK_{-v}$, we have that $\E[N_a(T)] = o(T^c), \ \forall c > 0$, where $N_a(T) := \sum_{v \in \gV_a} N_a^v(T)$.
\end{definition}
One obtains the following regret lower bound:
\begin{theorem}
\label{thm:generic-lower-bound}
    For any individually consistent policy $\Pi$,
    \begin{equation}
        \liminf_{T \rightarrow \infty} \frac{\E[R(T)]}{\log T}
        \geq \sum_{\substack{a \in \gK \\ \tilde{\Delta}_a > 0}} \frac{\tilde{\Delta}_a}{\inf_{P \in \gM_\sigma} \left\{ \KL(P_a, P) : \mu(P) - \mu(P_a) > \tilde{\Delta}_a \right\}},
    \end{equation}
    where $\gM_\sigma$ is the set of $\sigma$-sub-Gaussian distributions with mean $\mu$, and $\tilde{\Delta}_a = \min_{v \in \gV_{-a}} \Delta_a^v$.

   When $\gM_\sigma = \left\{ \gN(\mu, \sigma^2) : \mu \in \sR \right\}$, we obtain:
    \begin{equation}
        \liminf_{T \rightarrow \infty} \frac{\E[R(T)]}{\log T}
        \geq \sum_{\substack{a \in \gK \\ \tilde{\Delta}_a > 0}} \frac{2\sigma^2}{\tilde{\Delta}_a}.
    \end{equation}
\end{theorem}
The proof is immediate from the classic change-of-measure argument for cooperative multi-agent bandit setting~\cite{dubey2020coorperative,yang2022heterogeneous}.
Note that this asymptotic lower bound matches our asymptotic regret upper bound for both \textsc{Flooding} and \textsc{FwA} (Theorem~\ref{thm:ucb2}, Cor.~\ref{cor:ucb2}) up to some graph topology and arm distribution-dependent constants; see Appendix \ref{app:additional-lower} for a more detailed discussion.
\section{Experimental Results}
\label{sec:experiments}
In this section, we experimentally compare \textsc{FwA} to several existing algorithmic solutions.
The experiments were conducted on three random graph models with $N = 100$ nodes: the Erd\H{o}s-R\'{e}nyi model (ER)~\cite{ER1,ER2}, the Barab\'{a}si-Albert model (BA)~\cite{BA}, and the stochastic block model (SBM)~\cite{SBM}.
We use the following hyperparameters when generating the random graphs: for ER, the edge probability is set to $p_{ER} = 0.03$; for BA, the preferential attachment constant is set to $2$; for SBM, we consider 4 clusters, each with $N/4$ nodes, with intracluster and intercluster edge probabilities set to $0.3$ and $0.003$, respectively.

We set the total number of arms to $K=50$, and the number of arms per agent to be $k=20$.
We sample sets of size $k$ as arm sets for all the agents, uniformly at random.
The arm rewards follow Gaussian distributions, with the corresponding means uniformly sampled from $[0, 1]$ and fixed variance $\sigma^2 = 1$.
For all experiments, the overall arm distribution among the agents and the reward distributions are fixed.

We compare the baseline UCB with no cooperation between agents (baseline), \textsc{Flooding}, \textsc{Probabilistic Flooding (Prob. Flooding)}~\cite{oikonomou2023prob}, (uniform) \textsc{Gossip}, \textsc{IRS}, and our \textsc{FwA}.
For the~\textsc{Gossip} algorithm, we assume that each agent forwards messages to only one random neighbor at a time.
For \textsc{Prob. Flooding}, assuming that the learner has no prior knowledge of the communication network, we fix $q = 0.5$ to provide a fairer comparison.
We again emphasize that \textsc{FwA} does {\it not} require any tuning of hyperparameters other than the TTL.

All experiments are repeated $10$ times, with time horizon $T = 1000$.
We set TTL to be $\gamma = 4$, as our network instances' diameters are $8, 5, 7$ for ER, BA, and SBM, respectively, and we want our $\gamma$ to be strictly lower than all three values for meaningful results.
All codes were written in Python, and we made heavy use of the NetworkX package~\cite{networkx}.

\begin{figure*}[t]
     \centering  
     \vspace{-0.2cm}
     \hspace*{-1.7cm}
     \includegraphics[width=1.15\textwidth]{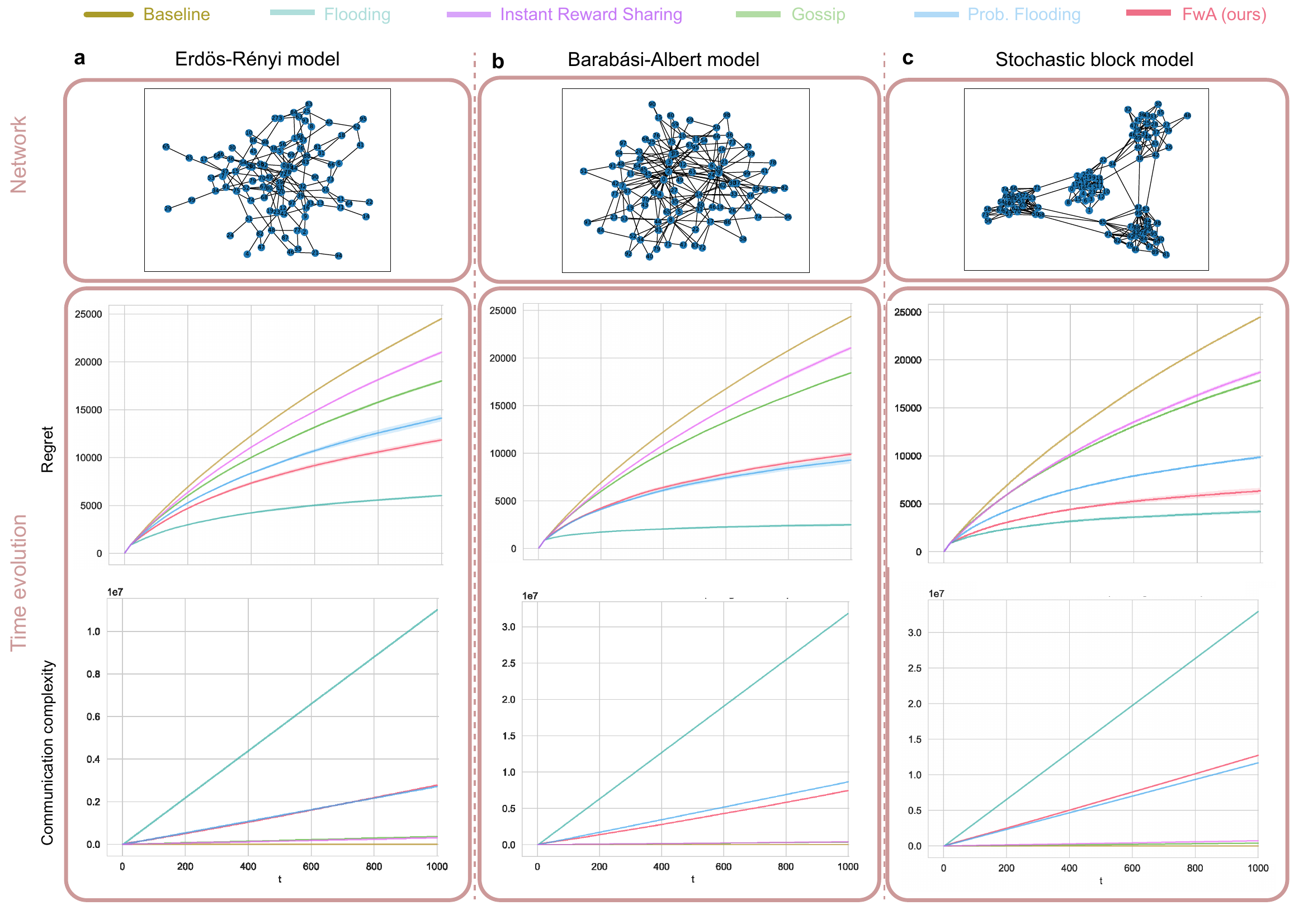}
     \caption{Comparing group regret and (cumulative) communication complexity across different topologies and protocols. Note that \textsc{FwA} gives a good trade-off between regret and communication complexity.}
     \label{fig:results1}
\end{figure*}

\subsection{Baseline comparison: Group Regret and Communication Complexity}
\label{sec:baseline}
We first compare the group regret and the communication complexity over time, i.e., the (cumulative) total number of messages sent, for all considered protocols.
The results are shown in Figure \ref{fig:results1}.
As expected, \textsc{Flooding} achieves the best regret out of the tested protocols, but its communication complexity is the worst.
Despite this, the important observation is that our \textsc{FwA} achieves second-best regret, arguably close to that achieved by \textsc{Flooding}, with a significantly reduced communication complexity.
Moreover, when compared to \textsc{Prob. Flooding}, \textsc{FwA} exhibits a better tradeoff between regret and communication: with a similar communication complexity, \textsc{FwA} achieves a better regret, or at least at par (for the BA model).
We also experimented with \textsc{Prob. Flooding} of other stopping probabilities (not shown here) and observed that they tend to show worse trade-offs between regret and communication complexity.
This shows that our proposed \textsc{FwA} protocol is a viable alternative to \textsc{Flooding} if one needs reduced communication complexity and good regret, uniformly across various network topologies.

\begin{figure}[!t]
    \centering
    \hspace*{-1.8cm}
    \resizebox{1.2\linewidth}{!}{
        \includegraphics{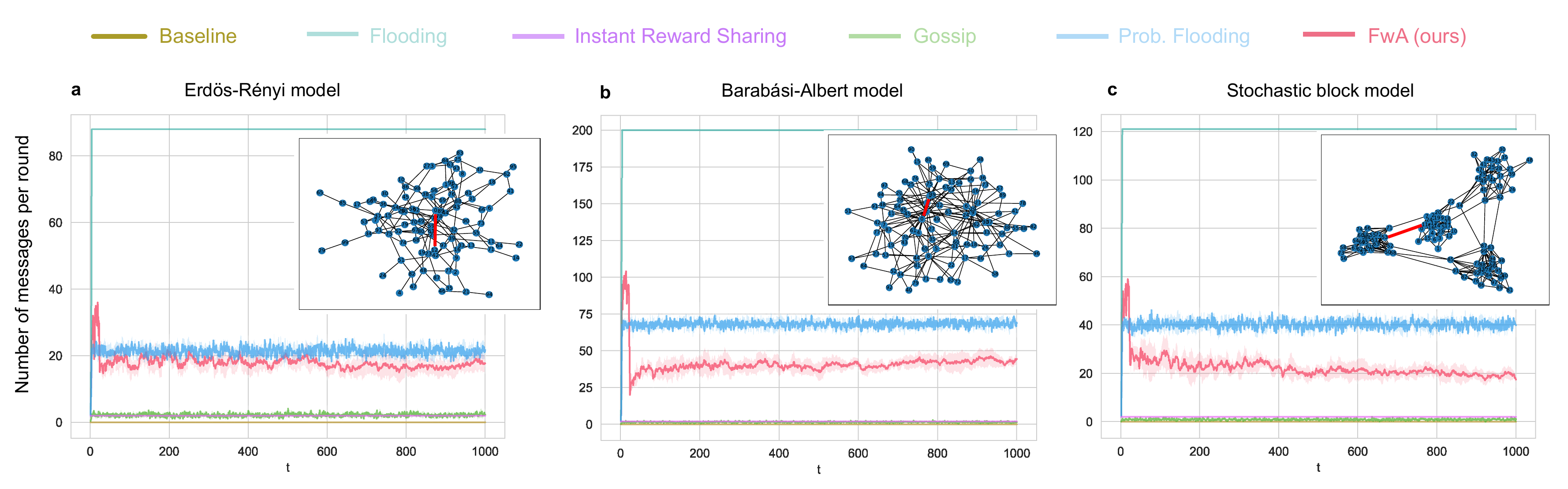}
    }
    \caption{\textsc{FwA} significantly decreases congestion on sparse network links. We find that, in comparison with other protocols, \textsc{FwA} results in a reduced number of messages sent over such a sparse link (highlighted in the networks).
    }
    \label{fig:link}
\end{figure}
\subsection{Link congestion}
In a setting where new messages are constantly produced by every agent (as each pulls an arm at each time step), one of the potential issues is link congestion caused by a large number of messages passing through bottleneck links.
This can lead to significantly decreased performance and undesirable latency effects - messages may be queued with limited memory in reliable link protocols, or automatically discarded in non-reliable link protocols once more messages are being sent than the link can handle.
In Figure~\ref{fig:link}, we visualize the number of messages sent over a particular link, again with TTL $\gamma=4$.
We chose a random link for ER and BA. For SBM, we chose a ``sparse'' link that connects two dense clusters.
Out of all the considered protocols, \textsc{FwA} results in the largest reduction of messages per round while providing good regret; specifically, for ER, BA, and SBM, \textsc{FwA} provides about $77\%, 80\%, 83\%$ reduction compared to \textsc{Flooding}, respectively.
We note that the reduction is larger even compared to \textsc{Prob. Flooding} - even though \textsc{Prob. Flooding} can occasionally have slightly better overall communication complexity (at worse regret).
This implies that our protocol exhibits significant benefits regarding individual network link congestion, which would help us avoid latency effects in real-life network applications.

One interesting observation is that \textsc{FwA} produces a spike in the number of messages in the early phase for all network topologies.
This is due to the design of the UCB algorithm; in the early phase, most of the agents are exploring the arms, and thus the messages are somewhat ``diverse''.
But as soon as the agents identify potentially best arms, the arm indices of the messages start to stabilize. They become less diverse, implying that from then on, absorption occurs more frequently under \textsc{FwA}.



\subsection{Dynamic networks}
\begin{figure}[!t]
    \centering
    \hspace*{-2cm}
    \resizebox{1.3\linewidth}{!}{
        \includegraphics{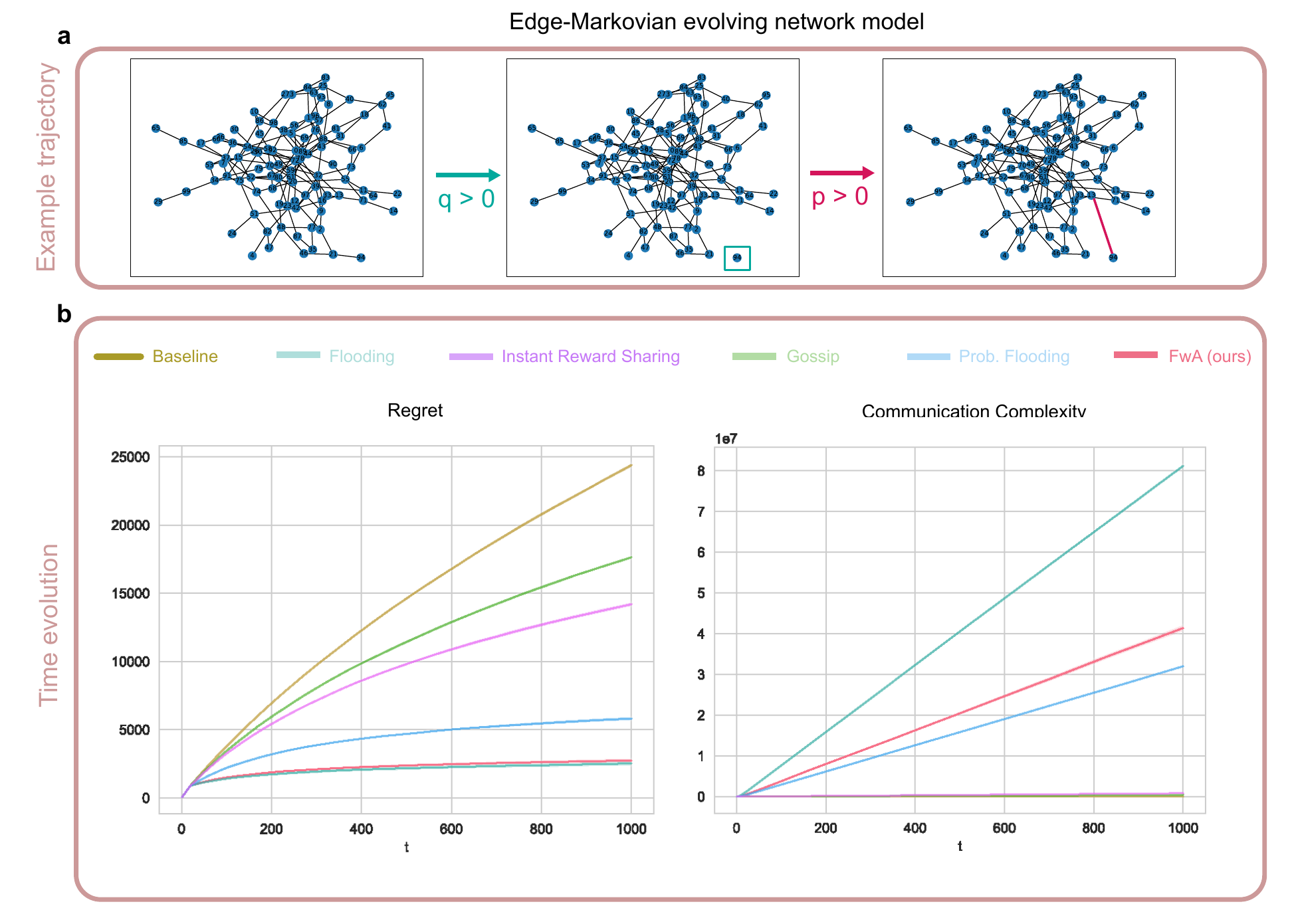}
    }
    \caption{Comparing group regret and (cumulative) communication complexity in a dynamic network setting. Note that \textsc{FwA} achieves the same regret as \textsc{Flooding}, with much lesser cumulative communication complexity.}
    \label{fig:dynamic}
\end{figure}
The advantage of our protocol is especially pronounced when we consider \emph{dynamically changing networks}, specifically, edge-Markovian networks~\cite{clementi2010markovian,clementi2011markovian,clementi2015markovian,clementi2016markovian}, where the network evolves as follows:
starting from an arbitrary initial graph $\gG_0 = (\gV, \gE_0)$, for $t \geq 1$, $\gG_t = (\gV, \gE_t)$ is stochastically determined as
\begin{equation}
    \sP[\{v, w\} \in \gE_t | \{v, w\} \not\in \gE_{t-1}] = p, \quad \sP[\{v, w\} \not\in \gE_t | \{v, w\} \in \gE_{t-1}] = q.
\end{equation}
$p$ and $q$ are often referred to as edge birth rate and edge death rate, respectively.
When $0 < p, q < 1$, it is well-known that, regardless of the initial graph $\gG_0$, the process converges to (stationary) Erd\H{o}s-R\'{e}nyi graph $\gG\left(\gV, \frac{p}{p + q} \right)$.
For our experiments, we start from the baseline ER graph and set $(p, q) = (0.01, 0.1)$. We plot an example trajectory in Figure~\ref{fig:dynamic}a.

The results are shown in Figure~\ref{fig:dynamic}b.
Observe how well our \textsc{FwA} protocol matches the regret of \textsc{Flooding}, with {\it strictly} better communication complexity. This trend is consistent across all considered networks, showing that \textsc{FwA} is the most effective out of the considered protocols in dynamic networks.
Similarly to the static case, we experimented with \textsc{Prob. Flooding} of other stopping probabilities (not shown here) and observed that they tend to show worse trade-offs between regret and communication complexity.
This suggests that the arm-dependent absorption mechanism of \textsc{FwA} implicitly regularizes the communication complexity in an efficient manner in dynamic networks, while minimizing the loss in regret.

\subsection{Tightness of Theoretical Regret Upper Bounds}
\label{sec:exp-delta}
\begin{figure}
     \centering
     \begin{subfigure}[!t]{0.49\textwidth}
         \centering
         \includegraphics[width=\textwidth]{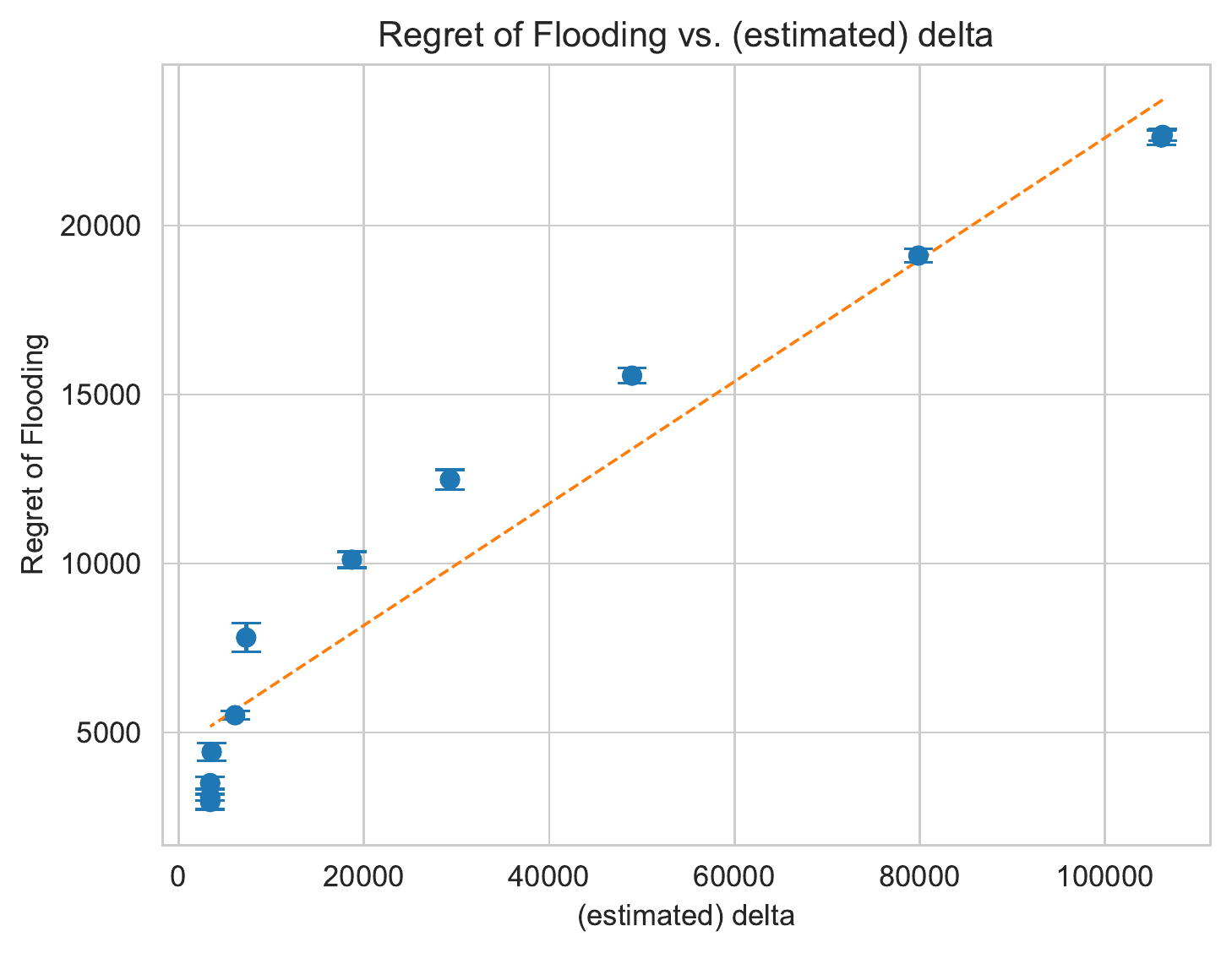}
         \caption{Regret of \textsc{Flooding} vs. $\delta^{Flooding}$. Orange line is the best linear fit ($R^2 = 0.9439$).}
         \label{fig:regret-delta-flooding}
     \end{subfigure}
     \hfill
     \begin{subfigure}[!t]{0.49\textwidth}
         \centering
         \includegraphics[width=\textwidth]{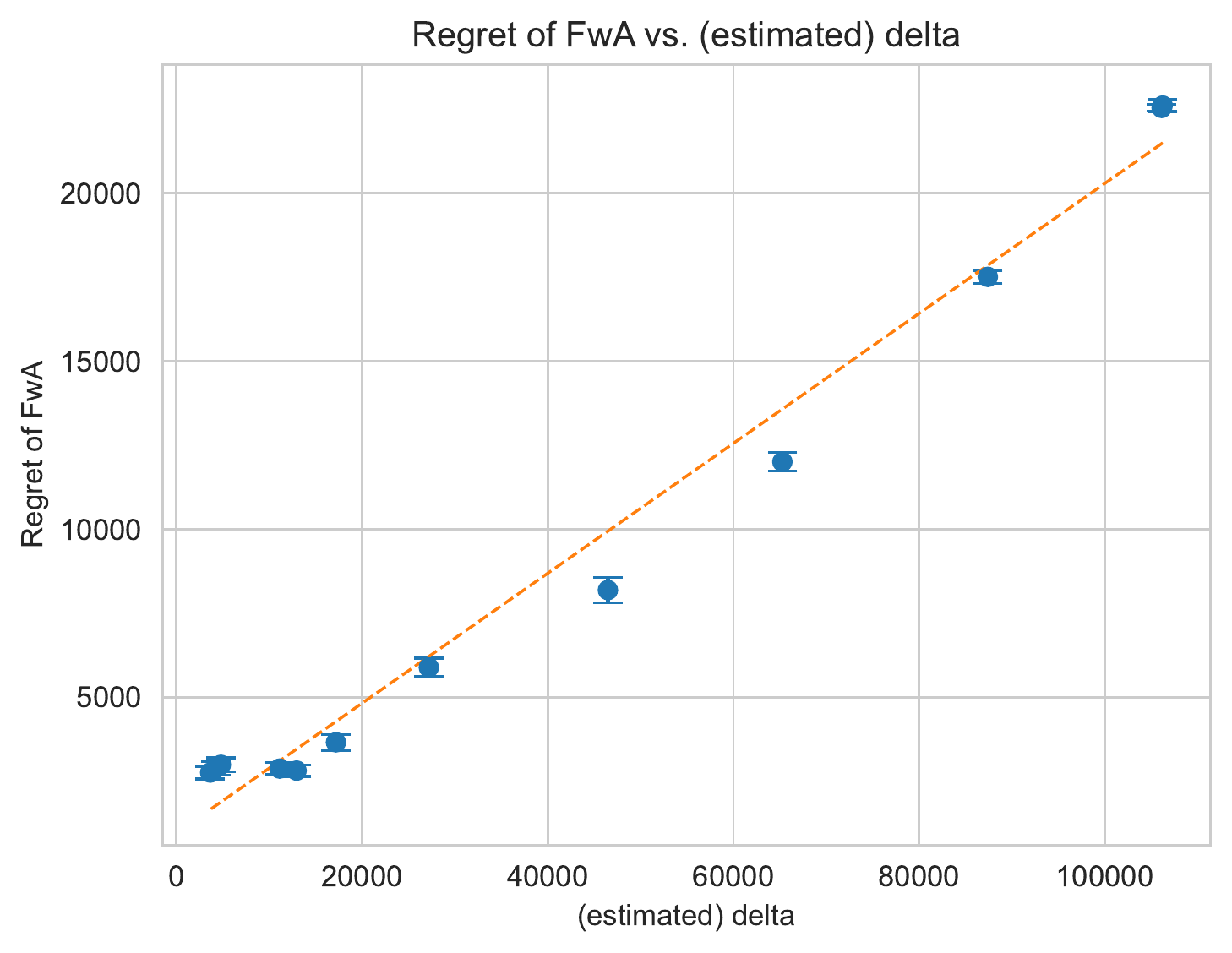}
         \caption{Regret of \textsc{FwA} vs. $\delta^{FwA}$. Orange line is the best linear fit ($R^2 = 0.9813$).}
         \label{fig:regret-delta-fwa}
     \end{subfigure}
    \caption{Experimental Results on $\delta$. Note that there is a strong linear correlation between the estimated $\delta^{Flooding}, \delta^{FwA}$ and the final resulting regrets of \textsc{Flooding}, \textsc{FwA}, respectively.}
    \label{fig:deltas}
\end{figure}

Recall that the theoretically derived regret bounds of \textsc{Flooding} and \textsc{FwA} (Theorem~\ref{thm:ucb1} and \ref{thm:ucb2}) depend on both network topology and arm distribution.
\textsc{Flooding} scales with $\delta^{Flooding} := \sum_{\substack{a \in \gK \\ \tilde{\Delta}_a > 0}} \frac{\theta([\gG^\gamma]_{-a})}{\tilde{\Delta}_a}$ and \textsc{FwA} scales with $\delta^{FwA} := \sum_{\substack{a \in \gK \\ \tilde{\Delta}_a > 0}} \frac{\theta\left([\gG_{(a, \vc)}^\gamma]_{-a}\right)}{\tilde{\Delta}_a}$.
To show that the theoretical bounds are tight and match well with practice, we perform an ablation study by varying the underlying edge density and seeing whether the aforementioned quantities scale well with the actual regrets.

We consider $\gG \sim G(n, p)$ of varying edge density $p$ and compare estimated $\delta$'s and regrets of \textsc{Flooding} and \textsc{FwA} under the same setting as in previous experiments.
Computing $\theta(\cdot)$'s requires computing the chromatic numbers, which is NP-hard.
We thus approximate it with the size of a greedy coloring of the considered graph
; for Erd\H{o}s-R\'{e}nyi graphs, the greedy coloring asymptotically results in twice the true chromatic number~\cite{frieze1997algorithmic}.

In Figure~\ref{fig:deltas}, we scatter plotted $(\delta, \text{Regret})$ along with a best linear fit for \textsc{Flooding} and \textsc{FwA}.
Indeed, it can be seen that the relationship is almost linear, with high $R^2$, showing that our regret bounds indeed reflect the regrets in practice.
There are some deviations from linearity, which we believe is due to small horizon length $T$ and inaccuracy in estimating $\delta$.
\section{Conclusion}
\label{sec:conclusion}

In this work, we described a novel setting for distributed multi-armed bandits, where agents communicate on an underlying network and do not all share the same arm set. We assume that each agent runs a UCB algorithm to identify their local best arm, and communicates the information they receive to their neighbors to minimize cumulative group regret. To deal with the very large communication complexity that however arises from using \textsc{Flooding} in our setting, we then introduced a new communication protocol for complex networks, \textsc{Flooding with Absorption (FwA)}. With \textsc{FwA}, agents forward information only if it pertains to an arm they themselves do not include in their arm set, whereas they absorb a message that gives information about one of their own arms.
We provided theoretical upper and lower regret bounds and showed experimentally that \textsc{FwA} incurs only minimal group regret performance loss compared to \textsc{Flooding} and even \textsc{Probabilistic Flooding}, while leading to a significantly improved communication complexity. In particular, we showed that \textsc{FwA} can reduce link congestion, which significantly improves upon simple heuristics such as probabilistic flooding. Our protocol is fully network-agnostic and hence does not need any fine-tuning, while still making use of the inherent heterogeneity of the problem instance. This makes it a very suitable choice for dynamically changing networks, or those that suffer from occasional message loss.
We believe that our work highlights the importance of integrating network topology and action heterogeneity in the design of distributed bandit algorithms, and provides an efficient way to connect bandit learning and network protocols.

\bibliography{_references}

\appendix

\section{Notations}
\label{app:notations}

\begin{table}[htbp]
\begin{center}
\small
\begin{tabular}{c c p{11cm} }
\toprule
\multicolumn{3}{l}{\bf Notations}\\
\hline
$\gG$ & & Communication network \\
$\gV$ & & Set of vertices in the graph, i.e., agents \\
$\gG[S]$ & & Induced subgraph over $S \subset \gV$ \\
$N=|\gV|$  & & The number of agents \\
$\gV_{a}$  & & Set of agents having arm $a$ \\
$\gV_{-a}$  & & Set of agents having arm $a$ as the suboptimal arm \\
$\gK$ & & The set of all arms in the network \\
$\gK_v$ & & The set of arms agent $v$ has access to \\
$\gK_{-v}$ & & The set of (locally) suboptimal arms agent $v$ has access to \\
$a_*^v$ & & The best local  arm for agent $v$ \\
$\mu_{a^v_*}$ & & Reward of best local arm for agent $v$\\
$\hat{\mu}_a^v(t)$ & & Average reward that agent $v$ has computed for arm $a$ at time $t$\\
$M_a^v(t)$ & & Number of observations of arm $a$ available to agent $v$ at time $t$\\
$N_a^v(t)$ & & Number of times agent $v$ pulls arm $a$ up to time $t$ \\
$N_a(t)$ & & Number of times all agents pull arm $a$ up to time $t$ \\
$\gamma$ & & Time-to-live of a message sent by an agent \\
$\gG^\gamma$ & & Graph power of $\gG$ of $\gamma$-th order:  graph over $\gV$ such that when $d_{\gG}(v,w) \leq \gamma$, $\{v,w\}$ is an edge \\
$T$ & & Time horizon \\
$R(T)$ & & Cumulative group regret over the time horizon \\
$\Delta_a^v$ & & Agent-specific suboptimality gap \\
\bottomrule
\end{tabular}
\end{center}
\end{table}

\section{Proof of Theorem \ref{thm:ucb1} -- Regret Upper Bound}
\label{proof:thm1}

The basic proof idea is to consider $\gG^\gamma$, which is equivalent to the collaboration network via flooding with TTL $\gamma$.
Then we consider a clique covering of $\gG^\gamma$, and upper bound the group regret of the full collaboration by that in which collaboration happens only intra-clique.
The deterministic delays between agents in each clique are precisely their distance, measured in the {\it original} communication network $\gG$.
We proceed similarly to \cite{yang2022heterogeneous} while filling in some missing details.

For each $a \in \gK$ and $v \in V_{-a}$ (i.e., $v$ contains $a$ as a {\it (locally) suboptimal} arm), define $B_a^v(t) := \{ a^v(t) = a \}$, which can be rewritten as
\begin{equation*}
    B_a^v(t) = \left\{ \hat{\mu}^v_{a'}(t) + \sqrt{\frac{\alpha \log t}{M^v_{a'}(t)}} < \hat{\mu}^v_a(t) + \sqrt{\frac{\alpha \log t}{M^v_a(t)}} \quad \forall a' \in \gK_v \setminus \{a\} \right\}.
\end{equation*}
Thus, this implies that the following event holds:
\begin{equation*}
    E^v_a(t) := \left\{ \hat{\mu}^v_{a^v_\star}(t) + \sqrt{\frac{\alpha \log t}{M^v_{a^v_\star}(t)}} < \hat{\mu}^v_a(t) + \sqrt{\frac{\alpha \log t}{M^v_a(t)}} \right\}.
\end{equation*}
i.e. $B_a^v(t) \subseteq E_a^v(t)$.

Next, for $\ell^v_a := \frac{4 \alpha \log T}{\left( \Delta_a^v \right)^2}$, define
\begin{equation}
    \tau^v_a := \min\left\{ t = 1, \cdots, T \ : \ M^v_a(t) > \ell^v_a \right\},
\end{equation}
i.e., $\tau^v_a$ is the first time at which agent $v$, and its neighbors, observe arm $a$ at least $\ell^v_a$ times.

Denoting $c_a^v(t) = \sqrt{\frac{\alpha \log t}{M^v_a(t)}}$, we have that~\cite{bubeck-thesis}
\begin{equation*}
    E_a^v(t) \subseteq
    \underbrace{\left\{ \hat{\mu}_{a_\star^v}^v(t) + c_{a_\star^v}^v(t) \leq \mu_{a_\star^v} \right\}}_{=A(t)} \cup
    \underbrace{\left\{ \hat{\mu}^v_a(t) \geq \mu_a + c_a^v(t) \right\}}_{=B(t)}
    \cup \left\{ M_a^v(t) < \ell^v_a \right\}.
\end{equation*}

Then the regret can be rewritten as follows~\cite{bubeck-thesis}:
\begin{equation*}
    R(T) = \E\left[ \sum_{a \in \gK} \sum_{v \in V_a} \Delta^v_a N_a^v(T) \right] \leq \underbrace{\sum_{a \in \gK} \sum_{v \in V_a} \Delta_a^v \E[N_a^v(\tau^v_a)]}_{(a)} + \underbrace{\sum_{a \in \gK} \sum_{v \in V_a} \Delta^v_a \sum_{t=1}^T \sP[E_a^v(t) \cap \{t > \tau_a^v\}]}_{(b)}.
\end{equation*}

(a) is bounded using clique covering and Abel transformation:
\begin{lemma}
\label{lem:(a)}
    For each $a \in \gK$ with $\gV_{-a} \not= \emptyset$, we have that
    \begin{equation*}
        \sum_{v \in V_a} \Delta_a^v \E[N_a^v(\tau_a^v)] \leq \frac{4 \alpha \log T}{\Delta_a^\gamma} + f_a(\gamma).
    \end{equation*}
\end{lemma}

(b) is bounded as follows:
\begin{lemma}
\label{lem:(b)}
    The following hold for each $v \in V$ and locally suboptimal $a \in \gK_v$:
    \begin{equation*}
        \sum_{t=1}^T \sP[E_t^v(a) \cap \{ t > \tau^v_a \}] \leq \left( \frac{\alpha + 1/2}{\alpha - 1/2} \right)^2 \frac{8(\gamma + 1)}{\log\frac{(\gamma + 1) (\alpha + 1/2)}{4\sigma^2}}.
    \end{equation*}
\end{lemma}

Combining both lemmas gives the desired statement.

\subsection{Proof of Lemma \ref{lem:(a)}}
Let $\gC_{-a}$ be a (vertex-disjoint) clique cover of $[\gG^\gamma]_{-a} = \gG^\gamma[\gV_{-a}]$, and let $C \in \gC_{-a}$.
Let $C = \{v_1, \cdots, v_c\}$ be such that $\Delta_a^{v_1} \geq \cdots \geq \Delta_a^{v_k} > 0$.

We fist have that $\E[N_a^{v_1}(\tau_a^{v_1})] \leq \E[M_a^{v_1}(\tau_a^{v_1})] \leq \frac{4\alpha \log T}{(\Delta_a^1)^2}$.
For each $k \in \{2, \cdots, c\}$,
\begin{align*}
    \sum_{j=1}^k \E[N_a^{v_j}(\tau_a^{v_j})] &\leq \E[N_a^{v_k}(\tau_a^{v_k})] + \sum_{j=1}^{k-1} \E[N_a^{v_j}(\tau_a^{v_j})] \\
    &\overset{(a)}{\leq} \E[M_a^{v_k}(\tau_a^{v_k})] + \sum_{j = 1}^{k-1} \sum_{\iota = 1}^\gamma \indicator[d_\gG(v_k, v_j) = \iota] \left( \E[N_a^{v_j}(\tau_a^{v_j})] - \E[N_a^{v_j}(\tau_a^{v_k} - \iota)] \right) \\
    &\overset{(b)}{\leq} \frac{4\alpha \log T}{(\Delta_a^{v_k})^2} + \sum_{j = 1}^{k-1} \sum_{\iota = 1}^\gamma \indicator[d_\gG(v_k, v_j) = \iota] \min\left( \gamma + \iota, \frac{4 \alpha \log T}{(\Delta_a^{v_j})^2} \right) \\
    &\leq \frac{4\alpha \log T}{(\Delta_a^{v_k})^2} + \sum_{j = 1}^{k-1} \min\left( 2\gamma, \frac{4 \alpha \log T}{(\Delta_a^{v_j})^2} \right).
\end{align*}
Here, $(a)$ follows from the simple decomposition of $M_a^{v_k}(\tau_a^{v_k})$:
\begin{align*}
    M_a^{v_k}(\tau_a^{v_k}) &= N_a^{v_k}(\tau_a^{v_k}) + \sum_{w \in \gV_a \setminus \{v_k\}} \sum_{\iota=1}^\gamma \indicator[d_\gG(v_k, w) = \iota] N_a^w(\tau_a^{v_k} - \iota) \\
    &\geq N_a^{v_k}(\tau_a^{v_k}) + \sum_{j=1}^{k-1} \sum_{\iota=1}^\gamma \indicator[d_\gG(v_k, v_j) = \iota] N_a^{v_j}(\tau_a^{v_k} - \iota).
\end{align*}
(b) follows from the following reasoning: first, we have that $\tau_a^{v_k} + \gamma \geq \tau^{v_j}_a$ as at time $\tau_a^{v_k} + \gamma$, any remaining agent $v_j$ should have at least $\ell_a^{v_k} \geq \ell_a^{v_j}$ observations of $a$, i.e., $\tau^{v_j}_a$ is already reached.
Thus, we have that 
\begin{equation*}
    \E[N_a^{v_j}(\tau_a^{v_j})] - \E[N_a^{v_j}(\tau_a^{v_k} - \iota)] \leq \E[N_a^{v_j}(\tau_a^{v_j})] \leq \frac{4 \alpha \log T}{(\Delta_a^{v_j})^2},
\end{equation*}
and
\begin{equation*}
    \E[N_a^{v_j}(\tau_a^{v_j})] - \E[N_a^{v_k}(\tau_a^{v_k} - \iota)] \leq
    \E[N_a^{v_j}(\tau_a^{v_k} + \gamma)] - \E[N_a^{v_j}(\tau_a^{v_k} - \iota)]
    \leq \iota + \gamma.
\end{equation*}


Recall that the group regret for our clique $C$ is $\sum_{k=1}^c \Delta_a^k \E[N_a^k(\tau_a^k)]$, where for simplicity we denote $\Delta_a^k := \Delta_a^{v_k}$.
The important observation is that to make the worst-case regret upper-bound, we must ``allocate'' the most number of pulls to the arms with the largest gap,
i.e.,
\begin{align*}
    &\sum_{k=1}^c \Delta_a^k \E[N_a^k(\tau_a^k)] \\
    &\leq \Delta_a^1 \left( \frac{4\alpha \log T}{(\Delta_a^1)^2} \right) + \sum_{k=2}^c \Delta_a^k \left\{ \left( \frac{4\alpha \log T}{(\Delta_a^k)^2} - \frac{4\alpha \log T}{(\Delta_a^{k-1})^2} \right) + \min\left( 2\gamma, \frac{4 \alpha \log T}{(\Delta_a^{k-1})^2}\right) \right\} \\
    &\overset{(*)}{=} \frac{4\alpha \log T}{\Delta_a^1} + \left( \Delta_a^c \frac{4\alpha \log T}{(\Delta_a^c)^2} - \Delta_a^1 \frac{4\alpha \log T}{(\Delta_a^1)^2} \right) + \sum_{k=1}^{c-1} \frac{4\alpha \log T}{(\Delta_a^k)^2} (\Delta_a^k - \Delta_a^{k+1}) + \sum_{k=2}^c \Delta_a^k \min\left( 2\gamma, \frac{4 \alpha \log T}{(\Delta_a^{k-1})^2}\right) \\
    &\leq \frac{4\alpha \log T}{\Delta_a^c} + \int_{\Delta_a^c}^{\Delta_a^1} \frac{4\alpha \log T}{z^2} dz + \sum_{k=2}^{c+1} \Delta_a^{k-1} \min\left( 2\gamma, \frac{4 \alpha \log T}{(\Delta_a^{k-1})^2}\right) \\
    &= \frac{8\alpha \log T}{\Delta_a^c} - \frac{4\alpha \log T}{\Delta_a^1} + \sum_{k=1}^c \Delta_a^k \min\left( 2\gamma, \frac{4 \alpha \log T}{(\Delta_a^k)^2}\right) \\
    &= 4\alpha\log T \left( \frac{2}{\min_{v \in C} \Delta_a^v} - \frac{1}{\max_{v \in C} \Delta_a^v} \right) + \sum_{k=1}^c \Delta_a^k \min\left( 2\gamma, \frac{4 \alpha \log T}{(\Delta_a^k)^2}\right),
\end{align*}
where $(*)$ follows from the Abel transformation.

Thus,
\begin{align*}
    &\sum_{v \in V_a} \Delta_a^v \E[N_a^v(\tau_a^v)] \\
    &= \min_{\gC_{-a}} \sum_{C \in \gC_{-a}} \sum_{v \in C} \Delta_a^v \E[N_a^v(\tau_a^v)] \\
    &\leq \min_{\gC_{-a}} \sum_{C \in \gC_{-a}} \left\{ 4\alpha\log T \left( \frac{2}{\min_{v \in C} \Delta_a^v} - \frac{1}{\max_{v \in C} \Delta_a^v} \right) \right\} + \sum_{v \in \gV_{-a}} \Delta_a^v \min\left( 2\gamma, \frac{4 \alpha \log T}{(\Delta_a^v)^2}\right) \\
    &= \frac{4\alpha \log T}{\Delta_a^\gamma} + f_a(\gamma).
\end{align*}

\subsection{Proof of Lemma \ref{lem:(b)}}
We start by noting that
\begin{equation*}
    \sP\left[ E_a^v(a) \cap \{ t > \tau^v_a \} \right] \leq \sP\left[A(t) \cap \{ t > \tau^v_a \}\right] + \sP[B(t) \cap \{ t > \tau^v_a \}].
\end{equation*}

We consider $\sP\left[A(t) \cap \{ t > \tau^v_a \}\right]$ first.
For simplicity, denote $M^v_\star(t) := M^v_{a_\star^v}(t)$.
The initialization phase of Algorithm \ref{alg:ucb-flooding} implies that
\begin{equation*}
    \sum_{\iota = 0}^\gamma \left| \left\{ w \in \gV_a : d_\gG(w, v) = \iota \right\} \right| \leq M_\star^v(t) \leq \sum_{\iota = 0}^\gamma (t - \iota) \left| \left\{ w \in \gV_a : d_\gG(w, v) = \iota \right\} \right|.
\end{equation*}
Denote $[a, b] := \{ \ceil{a}, \ceil{a}+1, \cdots, \floor{b}\}$, $\gN_\iota = \left\{ w \in \gV_a : d_\gG(w, v) = \iota \right\}$, and $N_\iota := \left| \gN_\iota \right|$.
Also, with a slight abuse of notation, here let us denote $X^u_{a_\star^v}(k)$ to be the reward received by agent $v$ when she pulls arm $a_\star$ for the $k$-th time.
Then,
\begin{align*}
    &\sP\left[A(t) \cap \{ t > \tau^v_a \}\right] \\
    &\leq \sP\left[ \exists \{s_\iota\}_{\iota \in [0, \gamma]}, s_\iota \in \left[ N_\iota, (t - \iota) N_\iota \right] \ \exists \{s_\iota(u)\}_{u \in \gN_\iota} : \sum_{u \in \gN_\iota} s_\iota(u) = s_\iota, \sum_{\iota=0}^\gamma s_\iota = s \right. \\
    &\quad \left. \ \text{ s.t. } \frac{1}{s} \sum_{\iota=0}^\gamma \sum_{u \in \gN_\iota} \sum_{k=1}^{s_\iota(u)} X^u_{a_\star^v}(k) + \sqrt{\frac{\alpha \log t}{s}} \leq \mu_\star^v \right] \\
    &\leq \sP\left[ \exists \{s_\iota\}_{\iota \in [0, \gamma]}, s_\iota \in \left[ N_\iota, (t - \iota) N_\iota \right] \ \exists \{s_\iota(u)\}_{u \in \gN_\iota} : \sum_{u \in \gN_\iota} s_\iota(u) = s_\iota, \sum_{\iota=0}^\gamma s_\iota = s \right. \\
    &\quad \left. \ \text{ s.t. } \sum_{\iota=0}^\gamma \sum_{u \in \gN_\iota} \sum_{k=1}^{s_\iota(u)} \left( X^u_{a_\star^v}(k) - \mu_\star^v \right) \leq - \sqrt{s \alpha \log t} \right] \\
    &\leq \sum_{\iota=0}^\gamma \sP\left[ \exists s_\iota \in \left[ N_\iota, (t - \iota) N_\iota \right] \ \text{ s.t. } \sum_{k=1}^{s_\iota} Y(k) \leq - \sqrt{s_\iota (\gamma+1) \alpha \log t} \right],
\end{align*}
where $Y(k) \overset{d}{=} X^u_{a_\star^v}(k) - \mu_\star^v $ is {\it i.i.d.} $\sigma$-subGaussian random variable with $\E[Y(k)] = 0$.

Before moving forward, we recall a maximal-type concentration result:
\begin{theorem}[Theorem 9.2 of \cite{lattimore2020bandit}]
    Let $Y_1, \cdots, Y_n$ be a sequence of independent $\sigma$-subGaussian random variables, and let $S_t = \sum_{s = 1}^t Y_s$.
    Then, for any $\varepsilon > 0$,
    \begin{equation}
        \sP\left[ \exists t \leq n : S_t \geq \varepsilon \right] \leq \exp\left( - \frac{\varepsilon^2}{2 n \sigma^2} \right).
    \end{equation}
\end{theorem}

Using the above concentration result as well as the peeling argument on a geometric grid \cite{bubeck-thesis,kolla2018collaborative}, we have that for any $\{\beta_\iota\}_{\iota \in [0, \gamma]}$ with $\beta_\iota \in \left( \frac{2\sigma^2}{(\gamma+1)\alpha}, 1 \right)$,

\begin{align*}
    &\sP[A(t)] \\
    &\leq \sum_{\iota=0}^\gamma \sum_{j=0}^{\frac{\log (t - \iota)}{\log(1/\beta_\iota)}} \sP\left[ \exists s_\iota \in \left[ N_\iota \beta_\iota^{j+1} (t - \iota), N_\iota \beta_\iota^j (t - \iota) \right] \ \text{ s.t. } \sum_{k=1}^{s_\iota} Y(k) \leq - \sqrt{s_\iota (\gamma+1) \alpha \log t} \right] \\
	&\leq \sum_{\iota=0}^\gamma \sum_{j=0}^{\frac{\log (t - \iota)}{\log(1/\beta_\iota)}} \sP\left[ \exists s_\iota \in \left[ N_\iota \beta_\iota^{j+1} (t - \iota), N_\iota \beta_\iota^j (t - \iota) \right] \ \text{ s.t. } \sum_{k=1}^{s_\iota} Y(k) \leq - \sqrt{ N_\iota \beta_\iota^{j+1} (t - \iota) (\gamma+1) \alpha \log t} \right] \\
	&\leq \sum_{\iota=0}^\gamma \sum_{j=0}^{\frac{\log (t - \iota)}{\log(1/\beta_\iota)}} \sP\left[ \exists s_\iota \in \left[ 1, N_\iota \beta_\iota^j (t - \iota) \right] \ \text{ s.t. } \sum_{k=1}^{s_\iota} Y(k) \leq - \sqrt{ N_\iota \beta_\iota^{j+1} (t - \iota) (\gamma+1) \alpha \log t } \right] \\
	&\leq \sum_{\iota=0}^\gamma \sum_{j=0}^{\frac{\log (t - \iota)}{\log(1/\beta_\iota)}} \exp\left( -\frac{N_\iota \beta_\iota^{j+1} (t - \iota) (\gamma+1) \alpha \log t}{2\sigma^2 N_\iota \beta_\iota^j (t - \iota)} \right) \\
	&= \sum_{\iota=0}^\gamma \left( 1+ \frac{\log (t - \iota)}{\log(1/\beta_\iota)} \right) t^{-\frac{\beta_\iota (\gamma+1) \alpha}{2\sigma^2}} \\
	&\leq \sum_{\iota=0}^\gamma \left( 1+ \frac{\log (t - \iota)}{\log(1/\beta_\iota)} \right) (t - \iota)^{-\frac{\beta_\iota (\gamma+1) \alpha}{2\sigma^2}}.
\end{align*}

One can show the same for $\sP[B(t)]$.
(Note how both of them do not depend on any graph theoretical quantities).

Then,
\begin{align*}
    \sum_{t=1}^T \sP[E_t^v(a) \cap \{ t > \tau^v_a \}] &\leq 2 \sum_{t=\tau^v_a+1}^T \sum_{\iota=0}^\gamma \left( 1+ \frac{\log (t - \iota)}{\log(1/\beta_\iota)} \right) (t - \iota)^{-\frac{\beta_\iota (\gamma+1) \alpha}{2\sigma^2}} \\
    &\leq 2 \sum_{\iota=0}^\gamma \sum_{t=\iota+1}^\infty \left( 1+ \frac{\log (t - \iota)}{\log(1/\beta_\iota)} \right) (t - \iota)^{-\frac{\beta_\iota (\gamma+1) \alpha}{2\sigma^2}} \\
    &\leq 2 \sum_{\iota=0}^\gamma \int_{\iota+1}^\infty \left( 1 + \frac{\log (t - \iota)}{\log(1/\beta_\iota)} \right) (t - \iota)^{-\frac{\beta_\iota (\gamma+1) \alpha}{2\sigma^2}} dt \\
    &= 2 \sum_{\iota=0}^\gamma \int_1^\infty \left( 1 + \frac{\log t}{\log(1/\beta_\iota)} \right) t^{-\frac{\beta_\iota (\gamma+1) \alpha}{2\sigma^2}} dt \\
    &= \frac{8(\gamma + 1)}{\left( \frac{\beta (\gamma+1) \alpha}{2\sigma^2} - 1 \right)^2 \log(1/\beta)},
\end{align*}
where we've set $\beta_\iota = \beta$ for all $\iota$'s.

Following \cite{bubeck-thesis}, we choose $\beta = \frac{4\sigma^2}{\gamma+1} \frac{1}{\alpha + \frac{1}{2}}$.
Then,
\begin{align*}
    \sum_{t=1}^T \sP[E_t^v(a) \cap \{ t > \tau^v_a \}] &\leq \left( \frac{\alpha + 1/2}{\alpha - 1/2} \right)^2 \frac{8(\gamma + 1)}{\log\frac{(\gamma + 1) (\alpha + 1/2)}{4\sigma^2}}.
\end{align*}

\section{Additional Discussions on the Theoretical Results}

\subsection{Main Technical Challenges 
in the Proof of Theorem~\ref{thm:ucb1} -- Regret Upper Bound}
\label{app:challenges}
One might ask whether it is possible to use a star decomposition-type argument for our setting similar to \cite{kolla2018collaborative}, which could give us an improvement from $\theta([\gG^\gamma]_{-a})$ to $\alpha([\gG^\gamma]_{-a})$.
We believe this is {\it not} possible, and the reason is as follows.

We note that the main technical challenge is that unlike the homogeneous settings \cite{madhushani2021regular,kolla2018collaborative}, the agents and the arms are intertwined: In a homogeneous setting, one could rewrite the regret as
\begin{equation*}
    R(T) = \sum_{v \in V} \sum_{a \in \gK} \Delta_a^{Kolla} \E[N_a^v(T)]
    = \sum_{a \in \gK} \Delta_a^{Kolla}  \sum_{v \in V} \E[N_a^v(T)],
\end{equation*}
i.e., the regret can be decomposed such that one only needs to upper bound the number of visitations of each agent $v$.
Thus with a star decomposition, for each star, it can be easily seen that the number of visitations of the leaf agents is precisely that of the center agent, allowing for us to further decompose $\sum_{v \in V} \E[N_a^v(T)]$ to sum of $\E[M_a^{v^{center}}(T)]$ over all center agents $v^{center}$.
This is at the core of dealing with more general homogeneous settings such as when there is a communication network \cite{kolla2018collaborative}, possibly with faults \cite{madhushani2021regular}.

Such a decomposition is {\it not} possible when the agents are heterogeneous, as the maximal suboptimality gap of $v$, $\Delta_a^v$, is agent-dependent.
To deal with such heterogeneity in the full information sharing (fully-connected graph) setting, Yang et al. \cite{yang2022heterogeneous} first ordered the agents according to $\Delta_a^v$, then bounded the {\it cumulative} number of times suboptimal arm $a$ is visited by agents $v_1, \cdots, v_k \in \gV_{-a}$ via the design of UCB algorithm.
Then, based on the intuition that the worst-case regret bound occurs when the arm $a$ with the highest $\Delta_a^v$ is visited at its maximum, the final regret bound is derived via the Abel transformation.
The fact that all agents have access to all other agents' information is crucial in this proof idea.
This can be seen from a very simple example; consider a situation in which agent $v_1$ has a very difficult problem (very small $\Delta_a^{v_1}$) but has lots of connections, and agent $v_2$ has a somewhat easy problem but has few connections.
In this case, from the collaboration, it may be that $v_1$ learns faster than $v_2$, which can impact the ordering of the agents, which in turn impacts the whole Abel transformation-based argument.
In a sense, our proof combines these two ideas.
We start with a clique decomposition, instead of the star decomposition, of the graph in order to upper-bound the group regret with the sum of regrets of each clique.
Then, we apply the Abel transformation-type argument to each clique.

Lastly, we remark that our results can also be easily extended to the setting where the agents asynchronously pull the arms, i.e., each agent $v$ pulls arms at every $\omega_v$ round, with $\omega_v \geq 1$.
It would be an interesting future direction if we could further reduce the communication cost in the asynchronous case based on ideas from recent works \cite{chen2023ondemand,yang2021global,yang2022heterogeneous}.

\subsection{Extending Theorem \ref{thm:generic-lower-bound} -- Regret Lower Bound}
\label{app:additional-lower}

In order to match the lower bound to the upper bound in terms of graph topology and arm distribution-dependent constant as well, there are two avenues, both of which are inspired by Kolla et al.~\cite{kolla2018collaborative}, who consider the homogeneous setting with a general graph.
On the one hand, it might be valuable to consider a Follow-Your-Leader-type policy, which would tighten the upper bound to match our lower bound.
However, it is unclear how to choose the leaders in our heterogeneous setting, let alone how to ensure that followers get relevant information on the arms.
Another way is to consider a more restricted class of policies, namely NAIC (non-altruistic \& individually consistent) policies, which would tighten the lower bound to match our upper bound.
Extending such notion to our setting of heterogeneous bandits over a graph {\it while} taking the communication protocol into account, e.g., whether we use the entire graph (\textsc{Flooding}) or we use part of the graph (\textsc{FwA}), and seeing whether we can match the lower bound up to the derived upper bounds, is another interesting future direction.
On a separate note, deriving a minimax lower bound, as done in Madhushani et al.~\cite{madhushani2021fault}, for our setting is also an interesting future direction.

\end{document}